\documentclass[preprint,12pt]{elsarticle}

\usepackage{amssymb}
\usepackage{amsmath}
\usepackage[colorlinks,urlcolor=blue,linkcolor=blue,citecolor=blue]{hyperref}
\usepackage{amsfonts}
\usepackage{algorithmic}
\usepackage{algorithm}
\usepackage{color, array}
\usepackage[caption=false,font=normalsize,labelfont=sf,textfont=sf]{subfig}
\usepackage{textcomp}
\usepackage{tabularx}
\usepackage{stfloats}
\usepackage{url}
\usepackage{verbatim}
\usepackage{graphicx}
\usepackage{xcolor}
\usepackage{cleveref}
\usepackage{geometry}
\usepackage{multirow}
\usepackage{array}
\usepackage{arydshln}
\graphicspath{{./}}

\definecolor{bluebox}{rgb}{0.000, 0.000, 1.000}   
\definecolor{greenbox}{rgb}{0.000, 1.000, 0.000}  
\definecolor{redbox}{rgb}{1.000, 0.000, 0.000}    
\definecolor{indigo}{HTML}{4B0082}
\definecolor{violet}{HTML}{8F00FF}

\begin{document}

\begin{frontmatter}



\title{Good Representation, Better Explanation: Role of Convolutional Neural Networks in Transformer-based Remote Sensing Image Captioning}

\author[label1]{Swadhin Das (s\_das@cs.iitr.ac.in)} 
\author[label2]{Saarthak Gupta (saarthak.gupta.mec22@itbhu.ac.in)}
\author[label3]{Kamal Kumar (2023nitsgr245@nitsri.ac.in)}
\author[label1]{Raksha Sharma (raksha.sharma@cs.iitr.ac.in)}
\affiliation[label1]{organization={Department of Computer Science and Engineering;\\ Indian Institute of Technology, Roorkee},
            addressline={Roorkee}, 
            city={Haridwar},
            postcode={247667}, 
            state={Uttarakhand},
            country={India}}
\affiliation[label2]{organization={Department of Mechanical Engineering;\\ Indian Institute of Technology,  BHU},
            addressline={Varanasi}, 
            city={Varanasi},
            postcode={221005}, 
            state={Uttar Pradesh},
            country={India}}
\affiliation[label3]{organization={Department of Information Technology;\\ National Institute of Technology, Srinagar},
    addressline={Srinagar}, 
    city={Srinagar},
    postcode={190006}, 
    state={Jammu and Kashmir},
    country={India}}

\begin{abstract}
Remote Sensing Image Captioning (RSIC) is the process of generating meaningful descriptions from remote sensing images. Recently, it has gained significant attention, with encoder-decoder models serving as the backbone for generating meaningful captions. The encoder extracts essential visual features from the input image, transforming them into a compact representation, while the decoder utilizes this representation to generate coherent textual descriptions. Recently, transformer-based models have gained significant popularity due to their ability to capture long-range dependencies and contextual information. The decoder has been well explored for text generation, whereas the encoder remains relatively unexplored. However, optimizing the encoder is crucial as it directly influences the richness of extracted features, which in turn affects the quality of generated captions. To address this gap, we systematically evaluate twelve different convolutional neural network (CNN) architectures within a transformer-based encoder framework to assess their effectiveness in RSIC. The evaluation consists of two stages: first, a numerical analysis categorizes CNNs into different clusters, based on their performance. The best performing CNNs are then subjected to human evaluation from a human-centric perspective by a human annotator. Additionally, we analyze the impact of different search strategies, namely greedy search and beam search, to ensure the best caption. The results highlight the critical role of encoder selection in improving captioning performance, demonstrating that specific CNN architectures significantly enhance the quality of generated descriptions for remote sensing images. By providing a detailed comparison of multiple encoders, this study offers valuable insights to guide advances in transformer-based image captioning models.
\end{abstract}

\begin{keyword}
Convolutional Neural Network (CNN) \sep Remote Sensing Image Captioning (RSIC) \sep Transformer-based Encoder \sep Transformer-based Decoder \sep Subjective Evaluation \sep K-Means Clustering

\end{keyword}

\begin{graphicalabstract}
\includegraphics[width=\textwidth,height=450px]{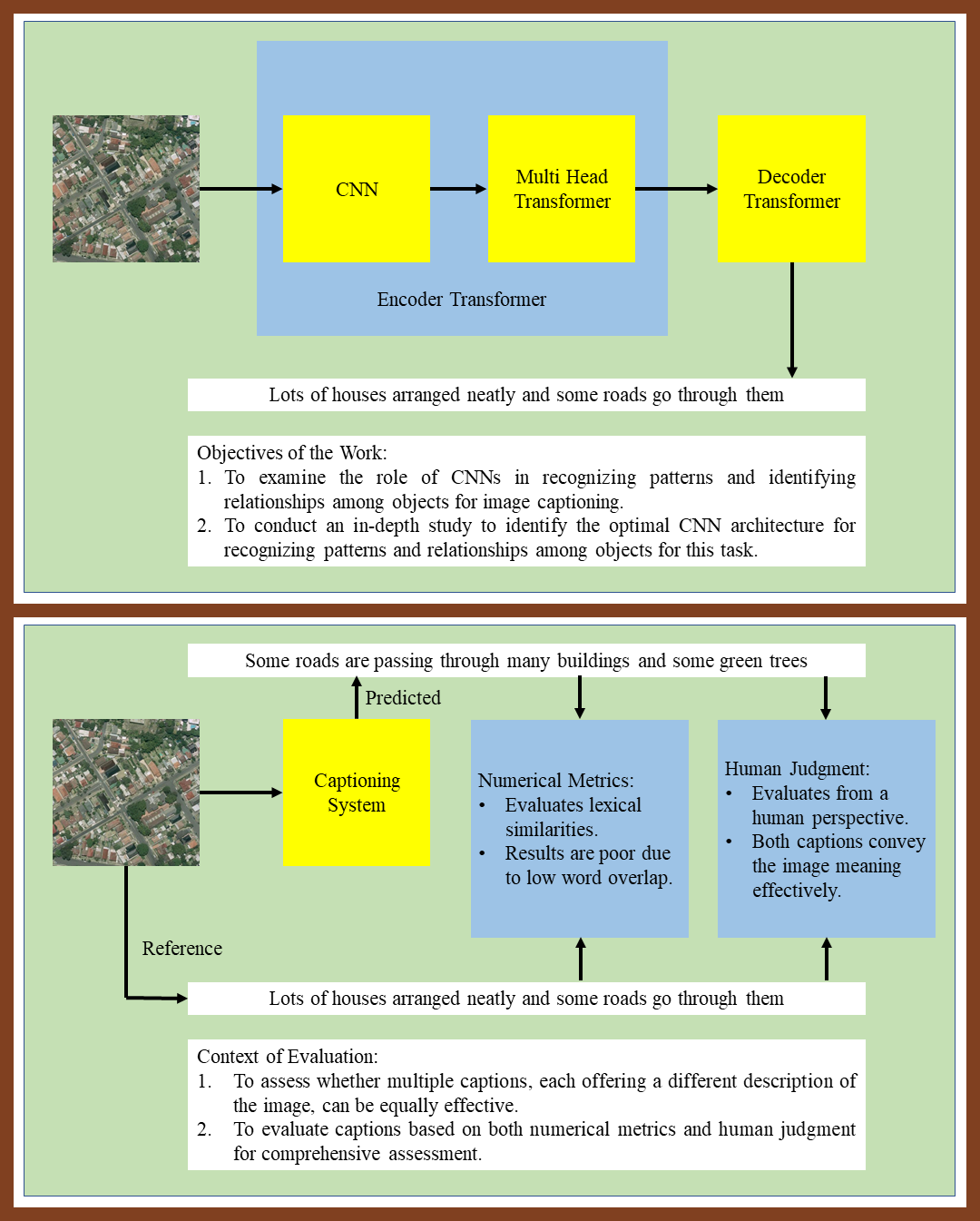}
\end{graphicalabstract}

\end{frontmatter}
\section{Introduction}
With the advancement of remote sensing technologies and machine learning-based methods, the demand for RSIC~\cite{lu2017exploring,das2024textgcn} is growing rapidly. It plays a crucial role in various fields, including environmental monitoring, urban planning, and disaster management, by providing automated textual descriptions of satellite images.  Unlike traditional machine learning tasks such as image classification, RSIC requires not only identifying objects in an image but also describing their relationships, spatial context, and scene composition in a meaningful way. While classification assigns predefined labels to images, RSIC generates natural language descriptions that capture fine-grained details, making it a more complex and context-dependent task. Since RSIC is highly domain specific, the quality of generated captions is heavily based on understanding the unique properties of remote sensing images. These images often contain diverse landscapes, multi-scale objects, and varying lighting conditions, requiring careful feature extraction and interpretation to produce accurate and informative captions.

The encoder-decoder-based model~\cite{qu2016deep} is one of the most widely used approaches in RSIC, where a suitable encoder (preferably a CNN) extracts visual features from the image and a decoder (such as RNN, LSTM, or GRU) generates captions based on the encoded representation. Although this approach produces reasonable results, it has several limitations. Traditional sequential models struggle with long-range dependencies, leading to issues such as loss of important contextual information, poor generalization of unseen images, and difficulty in capturing complex spatial relationships in remote sensing images. Furthermore, due to the high variability in satellite imagery, these models often do not generate semantically rich and contextually accurate captions. To address these challenges, attention-based models~\cite{zhang2023multi,li2021recurrent} were introduced, significantly improving RSIC performance. In this approach, the decoder selectively focuses on the relevant regions of the image while generating each word in the caption, ensuring better alignment between visual features and textual descriptions. This technique allows models to dynamically adjust their focus, improving the ability to describe objects, their relationships, and fine-grained scene details. As a result, many researchers adopted attention mechanisms in RSIC, achieving more accurate and contextually relevant captions. However, a major breakthrough came with the introduction of transformers, which use multi-head self-attention to process the entire input sequence simultaneously, capturing both local and global dependencies more effectively. With the success of transformers~\cite{vaswani2017attention} in natural language processing and vision tasks, RSIC research also started to shift toward transformer-based architectures, leveraging their ability to model complex relationships in remote sensing images. Despite these advancements, transformer-based RSIC~\cite{wu2024trtr} is still a relatively new concept and only a limited number of studies have explored its full potential. There remains significant scope for improving encoder design, attention mechanisms, and integration strategies to further enhance the accuracy and informativeness of generated captions.

The existing methods~\cite{hoxha2021novel,hoxha2020new,zhang2019multi} are well developed and produce strong results. However, most of the research has focused primarily on improving the decoder to generate high-quality captions, while the role of the encoder remains relatively underexplored. In RSIC, the encoder plays a crucial role, as the selection of an efficient encoder can significantly enhance the overall performance of the system. CNNs continue to be the preferred choice for image encoding, making it essential to thoroughly understand their effectiveness in this task. Given the diverse properties of different CNN architectures, a thorough evaluation is necessary to assess their effectiveness for RSIC. A systematic analysis of different CNNs as encoders will provide deeper insight into their impact on captioning performance, ultimately contributing to the generation of more accurate and informative descriptions.

To overcome the above challenges, we examined twelve different CNNs within a transformer-based image encoder framework for RSIC. We analyze these widely used CNN architectures, evaluating their ability to extract meaningful visual features that enhance caption generation. To assess their behavior, we performed experiments from multiple perspectives. First, a numerical analysis was performed, aggregating CNNs into three groups:~\emph{Good},~\emph{Medium}, and~\emph{Bad}. The~\emph{Good} CNNs were further evaluated by a human annotator to assess the quality of their generated captions from a human perspective. The contributions of this paper are as follows.  
\begin{itemize}  
    \item We conducted a thorough numerical evaluation of CNNs and identified the top performing architectures.  
    \item We performed a subjective evaluation of the selected CNNs.    
    \item We strengthened the analysis by conducting a classification task using CNNs in our image dataset.    
    \item We demonstrated the impact of multi-head transformers in RSIC through ablation studies.  
\end{itemize} 
The rest of the paper is organized as follows:~\Cref{sec_relwork} reviews related work, discussing previous research on RSIC and the role of different encoder-decoder architectures.~\Cref{sec_methodology} details the proposed methodology, including the integration of CNN-based feature extraction within the transformer-based encoder framework.~\Cref{sec_CNN} presents an overview of the CNN architectures evaluated in this study.~\Cref{sec_exp_setup} describes the experimental setup, including datasets and evaluation metrics.~\Cref{sec_experimental} provides an in-depth analysis of numerical results, subjective evaluations and error patterns. Finally,~\Cref{sec_conclusion} concludes the study with key findings and future research directions.
\section{Related Work}
\label{sec_relwork}

Recent advancements in deep learning have significantly transformed remote sensing image captioning (RSIC), which focuses on generating descriptive text from satellite or aerial images. Although early efforts relied on rule-based techniques and handcrafted features, the introduction of data-driven architectures enabled feature extraction and caption generation automation. These developments positioned RSIC as a crucial task for interpreting complex visual scenes and contributed to a broader trend of applying image captioning techniques to high-resolution remote sensing data.

A significant milestone in RSIC was the adoption of encoder-decoder architectures, which allowed models to be trained end-to-end. These frameworks typically employ a CNN to encode the image and a recurrent neural network (RNN) to generate corresponding captions. Qu et al.,~[2016]~\cite{qu2016deep} proposed a deep multimodal network that fuses visual and textual embeddings, laying the foundation for future work. Lu et al.,~[2017]~\cite{lu2017exploring} addressed limitations in early datasets by introducing the RSICD dataset to enhance diversity and coverage. Li et al.,~[2019]~\cite{li2019vision} further improved these frameworks by introducing a two-level attention mechanism, while Zhang et al.,~[2019]~\cite{zhang2019multi} used multiscale cropping to enhance spatial feature granularity. Hoxha et al.,~[2020]~\cite{hoxha2020new} refined decoding with beam search and retrieval-based enhancements. Hoxha et al.,~[2020]~\cite{hoxha2020toward} incorporated caption-guided retrieval using CNN–RNN models.

Attention mechanisms were introduced within the encoder-decoder pipeline to capture context and focus on relevant image regions during caption generation. These mechanisms dynamically weight different image features depending on the evolving sentence context. Xu et al.,~[2015]~\cite{xu2015show} pioneered this approach, which inspired attention-based enhancements in RSIC. Sumbul et al.,~[2020]~\cite{sumbul2020sd} proposed a summarization-driven attention model to improve semantic relevance and sentence structure. Li et al.,~[2021]~\cite{li2021recurrent} introduced the RASG framework, which uses recurrent connections to refine attention vectors. Wang et al.,~[2022]~\cite{wang2022glcm} combined global and local visual cues to improve feature discrimination and caption accuracy.

Although attention mechanisms improved context modeling, challenges remained in capturing long-range dependencies and maintaining linguistic coherence. Transformer-based architectures addressed these issues by introducing self-attention and parallel processing capabilities. Originally proposed by Vaswani et al.,~[2017]~\cite{vaswani2017attention}, transformers have since been widely adopted in RSIC. Liu et al.,~[2022]~\cite{liu2022remote} developed MLAT, a multilayer transformer that aggregates multiscale features for a richer representation. Meng et al.,~[2023]~\cite{meng2023prior} designed the PKG transformer with graph neural networks to model object relations more effectively. Zhang et al.,~[2023]~\cite{zhang2023multi} introduced a stair-step attention model reinforced by CIDEr optimization. Lin et al.,~[2024]~\cite{lin2024clip} proposed a CLIP-based transformer model with visual grid features and random text masking to enhance remote sensing image captioning. Wu et al.,~[2024]~\cite{wu2024trtr} leveraged dual-transformer modules with Swin blocks for enhanced multiscale extraction. Meng et al.,~[2025]~\cite{meng2025rsic} proposed RSIC-GMamba, a Transformer-enhanced Mamba framework that integrates genetic operations and self-attention to capture multiscale visual context for RSIC.

Despite the effectiveness of these methods, most RSIC research has focused on decoder architectures, often overlooking the influence of the encoder. However, the encoder is equally critical for producing meaningful representations that drive caption quality. Das et al.,~[2024]~\cite{das2024unveiling} investigated this by comparing eight CNN-based encoders using an LSTM decoder. Their performance-based grouping and subjective evaluation identified ResNet as the best encoder in their setup. Building upon this motivation, the present work expands the scope to twelve encoders and adopts a modern architecture composed of a transformer-based encoder and a GPT-2 decoder. This unified configuration enables a more comprehensive and current evaluation of encoder contributions to RSIC performance.

\section{Methodology}
\label{sec_methodology}
\begin{figure*}[!ht]
    \centering
    \includegraphics[width=0.95\textwidth]{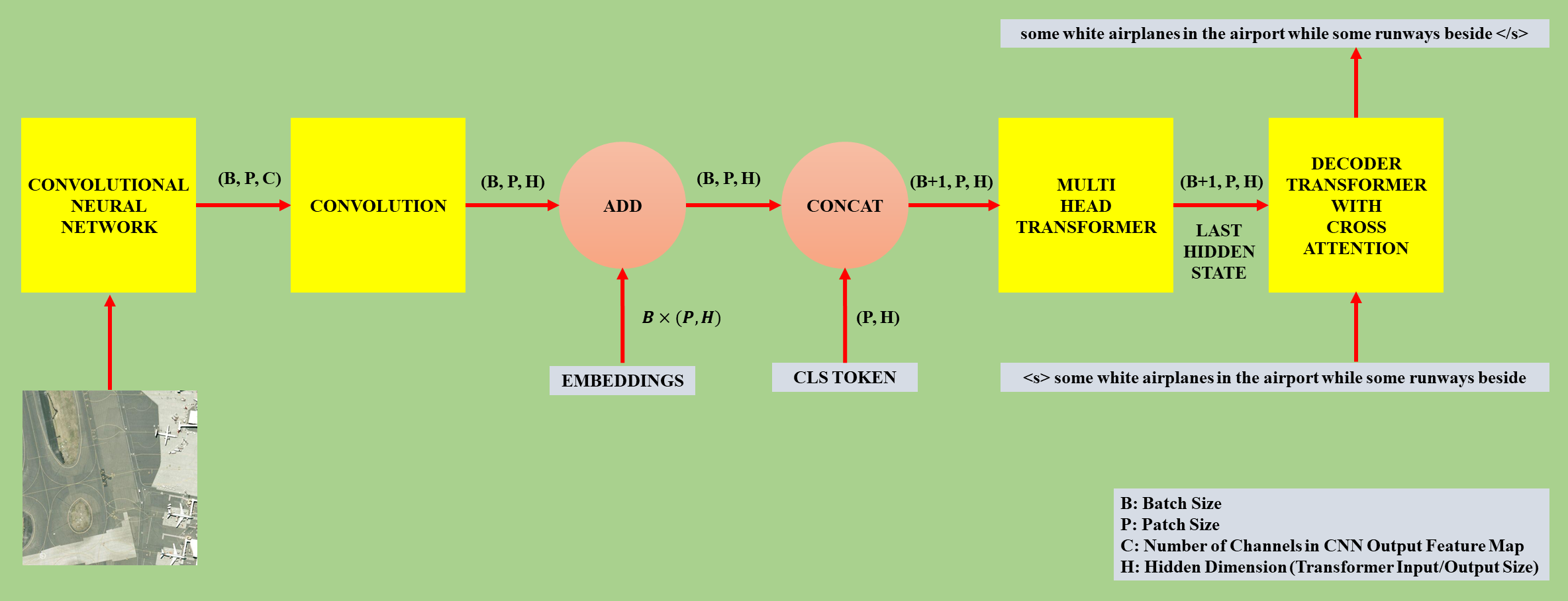}
    \caption{Architecture of the model}
    \label{architecture}
\end{figure*}
~\Cref{architecture} illustrates the workflow of our model. In this work, we adopt an encoder-decoder-based architecture to generate captions from remote sensing images. The encoder consists of a hybrid module that includes a CNN-based feature extractor followed by a multi-layer transformer. This joint design integrates local spatial representation learning (via CNN) with global context modeling (via self-attention), forming a unified encoder pipeline. First, an input image is processed by the CNN, and the resulting output is permuted into the shape (batch size, patch size, feature size). A $1\times1$ convolution layer is then applied to standardize the~\emph{feature size}, as different CNN architectures produce outputs of varying dimensions. Next, we add~\emph{patch embeddings}, which allows the model to divide the image into smaller regions, thus capturing both local and global features more effectively while maintaining positional information. Unlike classification-based models, we do not include a~\emph{CLS token}, as it is not required for our captioning objective. The resulting sequence of embedded features is passed through the transformer encoder layers repetitively. Finally, the encoded features are forwarded to the transformer-based decoder to generate the caption.

In our approach, we employ GPT-2 as the transformer-based decoder, enhanced with cross-attention mechanisms to generate captions from remote sensing images. GPT-2 is a substantial transformer-based language model trained on a diverse corpus of text data, excelling at producing coherent and contextually relevant text sequences, making it well-suited for captioning tasks. Within GPT-2, the self-attention mechanism enables the model to consider all positions in the input sequence simultaneously when making predictions, effectively capturing long-range dependencies and relationships between elements. To prevent the model from attending to future tokens during training or generation, masked self-attention is employed. This ensures that the model focuses only on the current and previous tokens, maintaining the autoregressive nature of text generation. In our work, we integrate a cross-attention layer that aligns the encoded image features with the text generation process. This mechanism allows the decoder to focus on relevant visual information while predicting each token, effectively bridging the gap between visual and textual modalities. The cross-attention mechanism operates by using the decoder's queries to attend to the encoder's key-value pairs, enabling the model to incorporate pertinent image features into the caption generation process. By leveraging GPT-2's pre-trained language understanding capabilities and incorporating both masked self-attention and cross-attention mechanisms, our model generates semantically meaningful captions that accurately describe remote sensing images.  

Each sentence is appended with a start token \(\langle s\rangle\) and an end token \(\langle/s\rangle\) to mark the beginning and end of a sentence. We use GPT2Tokenizer (fast version) to tokenize sentences. Both the tokenizer and the decoder model are fine-tuned on the training caption dataset, as the captions are highly domain-specific. Fine-tuning ensures that the model learns the specialized vocabulary and structure required to accurately describe remote sensing images.  

The training phase differs significantly from the RNN- or LSTM-based decoders. Here, we pass the entire sequence of sentence tokens, including the start token but excluding the end token, and compare the output with the sequence excluding the start token. The main objective is to train the model to predict the next token for each input token.  

The sentence generation process follows a straightforward approach. It begins with the start token as the input sequence. The image feature extraction process remains the same. Using these two inputs, the model predicts the next word and appends it to the input sequence. This iterative process continues until the model generates the end token.
\section{Types of CNN}
\label{sec_CNN}
\begin{table*}[!ht]
\centering
\caption{Computational Complexities of Various CNN Backbones (in Billions)}
\label{model_stats}
\resizebox{\textwidth}{!}{
\begin{tabular}{|c|c|c|c|c|c|c|c|c|c|c|c|c|c|c|c|}
\hline
Metric & ResNet & WideResNet & ResNext & RegNet & VGGNet & DenseNet & AlexNet & GoogleNet & InceptionNet & MobileNetV2 & MobileNetV3 & SqueezeNet & ShuffleNet & MNASNet & ConvNext \\
\hline
FLOPs & 27.7041 & 50.1723 & 35.6666 & 68.2882 & 43.5181 & 13.2761 & 4.5302 & 7.4518 & 37.6403 & 5.0834 & 4.8788 & 16.5252 & 5.6930 & 5.5508 & 73.3968 \\
\#PARAMS & 0.0668 & 0.1335 & 0.0901 & 0.1144 & 0.0275 & 0.0267 & 0.0098 & 0.0135 & 0.0169 & 0.0103 & 0.0108 & 0.0092 & 0.0140 & 0.0131 & 0.2045 \\
\hline
\end{tabular}}
\end{table*}
Our work incorporates twelve widely recognized CNN architectures within the transformer-based encoder framework. Although there are multiple variants for each CNN type, we selected the most recent versions to maintain a fair and consistent comparison. In the following, we briefly describe the motivation behind choosing each CNN architecture and the key design principles that define their functionality.
\begin{enumerate}
    \item\textbf{ResNet~\cite{he2016deep}:} ResNet addresses the problem of vanishing gradients in deep neural networks by implementing residual connections. These connections allow models to learn identity mappings, facilitating stable gradient flow and efficient feature learning. Comprising multiple residual blocks, ResNet enables the training of very deep networks without performance degradation. We used ResNet with $152$ layers in our work.
    \item\textbf{Wide ResNet~\cite{zagoruyko2016wide}:} Wide ResNet (WRN) is a variant of ResNet developed through an experimental study of its architecture. It enhances training efficiency and performance by reducing the depth while increasing the width of residual networks. This modification enhances feature representation and mitigates the degradation problem in deep networks. 
    \item\textbf{ResNext~\cite{xie2017aggregated}:} ResNext is a deep learning architecture that expands on ResNet by introducing~\emph{cardinality}, the number of transformation sets in each block. It uses a simple, repeatable multibranch design with minimal hyperparameters, enhancing model flexibility and performance. This approach improves feature learning while maintaining efficiency.
    \item\textbf{RegNet~\cite{radosavovic2020designing}:} RegNet defines a design space of simple, regular networks with quantized linear parameterization of widths and depths, allowing for scalable and efficient models. This approach optimizes performance across different computational budgets, balancing accuracy and efficiency.
    \item\textbf{VGGNet~\cite{simonyan2014very}:} VGGNet is one of the earliest and most widely used CNNs for tasks like image captioning. It is known for its substantial number of parameters. We used VGGNet with $19$ layers in our work.
    \item\textbf{DenseNet~\cite{huang2017densely}:} DenseNet connects each layer to every other layer in a feed-forward fashion. This dense connectivity improves the flow of the gradient and facilitates better feature reuse. We used DenseNet with $201$ layers in our work.
    \item\textbf{AlexNet~\cite{krizhevsky2012imagenet}:} AlexNet was among the first deep CNNs and is capable of performing parallel processing using two GPUs.
    \item\textbf{GoogleNet~\cite{szegedy2015going}:} GoogleNet is the first version of InceptionNet. Using inception modules, it integrates multiple parallel convolutional and pooling operations with varying kernel sizes.
    \item\textbf{InceptionNet~\cite{szegedy2016rethinking}:} InceptionNet is an advanced version of GoogleNet. Improves the architecture. For our research, we used InceptionNet version three in our work.
    \item\textbf{MobileNetV2~\cite{sandler2018MobileNetV2}:} MobileNetV2 introduces a mobile architecture that improves performance across various tasks and model sizes. It uses an inverted residual structure with lightweight depthwise convolutions and eliminates non-linearities in narrow layers to enhance representational power. 
    \item\textbf{MobileNetV3~\cite{howard2019searching}:} MobileNetV3 improves on previous MobileNet architectures by using hardware-aware network architecture search (NAS) and the NetAdapt algorithm, optimizing models for mobile devices. With two variants, MobileNetV3-Large and MobileNetV3-Small, the architecture achieves superior performance in classification, detection, and segmentation tasks, offering higher accuracy and reduced latency compared to MobileNetV2. We used MobileNetV3 of a large variant in our work.
    \item\textbf{ConvNext~\cite{liu2022convnet}:} ConvNext is an advanced CNN that adopts key design principles of vision transformers. By integrating techniques such as depthwise convolutions and layer normalization, it enhances traditional CNN architectures, offering better performance on image classification tasks while maintaining computational efficiency. This approach positions ConvNext as a strong alternative to vision transformers.
\end{enumerate}

To better contextualize the computational footprint, we categorize CNN backbones into three groups based on their parameter sizes, as summarized in~\Cref{model_stats}: High-Capacity, Mid-Range, and Lightweight Architectures. High-capacity models (such as ConvNext, WideResNet, RegNet, and ResNext) offer strong representational power and demand significantly higher computational resources. Mid-range networks (such as ResNet, VGGNet, and DenseNet) provide a balance between efficiency and performance, making them suitable for moderately constrained setups. Lightweight models (such as AlexNet, GoogleNet, InceptionNet, MobileNetV2, MobileNetV3, ShuffleNet, MNASNet, and SqueezeNet) are designed for environments with limited resources and offer compact architectures with minimal parameter overhead.
\section{Experimental Setup}
\label{sec_exp_setup}
\begin{table*}[!ht]
\centering
\caption{Results of Various CNNs on the  SYDNEY Dataset with Greedy Search}
\label{SYDNEY_greedy}
\resizebox{\textwidth}{!}{
\begin{tabular}{|c|c|c|c|c|c|c|c|c|}
\hline
CNN & BLEU-1 & BLEU-2 & BLEU-3 & BLEU-4 & METEOR & ROUGE-L & CIDEr & Cluster\\
\hline
ResNet & 0.7480 & 0.6484 & 0.5719 & 0.5039 & 0.3630 & 0.6723 & 2.0973 & Good\\
Wide ResNet & 0.7131 & 0.6295 & 0.5621 & 0.5063 & 0.3813 & 0.6636 & 2.1904 & Medium\\
ResNext & 0.7609 & 0.6658 & 0.5815 & 0.5078 & 0.3760 & 0.6852 & 2.1877 & Good\\
RegNet & 0.7105 & 0.6087 & 0.5367 & 0.4778 & 0.3492 & 0.6395 & 1.8811 & Bad\\
VGGNet & 0.7118 & 0.6264 & 0.5592 & 0.4991 & 0.3744 & 0.6739 & 2.0582 & Medium\\
DenseNet & 0.7250 & 0.6331 & 0.5597 & 0.4977 & 0.3548 & 0.6527 & 1.8792 & Bad\\
AlexNet & 0.7239 & 0.6314 & 0.5517 & 0.4881 & 0.3569 & 0.6738 & 2.2170 & Medium\\
GoogleNet & 0.7391 & 0.6419 & 0.5634 & 0.4971 & 0.3552 & 0.6681 & 2.0574 & Medium\\
InceptionNet & 0.7567 & 0.6667 & 0.5926 & 0.5299 & 0.3760 & 0.6767 & 2.1791 & Good\\
MobileNetV2 & 0.6986 & 0.5919 & 0.5161 & 0.4528 & 0.3328 & 0.6227 & 1.7561 & Bad\\
MobileNetV3 & 0.7461 & 0.6528 & 0.5769 & 0.5122 & 0.3465 & 0.6623 & 2.0603 & Medium\\
ConvNext & 0.7997 & 0.6844 & 0.6325 & 0.5694 & 0.4073 & 0.7349 & 2.4945 & Good\\
\hline
\end{tabular}}
\end{table*}
\begin{table*}[!ht]
\centering
\caption{Results of Various CNNs on the  SYDNEY Dataset with Beam Search}
\label{SYDNEY_beam}
\resizebox{\textwidth}{!}{
\begin{tabular}{|c|c|c|c|c|c|c|c|c|}
\hline
CNN & BLEU-1 & BLEU-2 & BLEU-3 & BLEU-4 & METEOR & ROUGE-L & CIDEr & Cluster\\
\hline
ResNet & 0.7578 & 0.6663 & 0.5906 & 0.5228 & 0.3729 & 0.6883 & 2.1498 & Good\\
Wide ResNet & 0.7202 & 0.6385 & 0.5721 & 0.5139 & 0.3635 & 0.6380 & 2.1634 & Medium\\
ResNext & 0.7298 & 0.6386 & 0.5628 & 0.4964 & 0.3589 & 0.6559 & 2.1403 & Medium\\
RegNet & 0.7060 & 0.6125 & 0.5408 & 0.4781 & 0.3490 & 0.6422 & 1.9920 & Bad\\
VGGNet & 0.7267 & 0.6387 & 0.5745 & 0.5206 & 0.3796 & 0.6860 & 2.3559 & Good\\
DenseNet & 0.7359 & 0.6465 & 0.5730 & 0.5107 & 0.3557 & 0.6706 & 2.0070 & Medium\\
AlexNet & 0.6879 & 0.6027 & 0.5315 & 0.4786 & 0.3266 & 0.6174 & 2.1447 & Bad\\
GoogleNet & 0.7204 & 0.6293 & 0.5515 & 0.4855 & 0.3627 & 0.6570 & 2.0371 & Medium\\
InceptionNet & 0.7384 & 0.6538 & 0.5859 & 0.5281 & 0.3610 & 0.6654 & 2.2171 & Good\\
MobileNetV2 & 0.7044 & 0.6039 & 0.5284 & 0.4647 & 0.3419 & 0.6395 & 1.7789 & Bad\\
MobileNetV3 & 0.7055 & 0.6139 & 0.5417 & 0.4789 & 0.3490 & 0.6451 & 1.9865 & Bad\\
ConvNext & 0.7762 & 0.6742 & 0.6255 & 0.5513 & 0.3955 & 0.7281 & 2.4176 & Good\\
\hline
\end{tabular}}
\end{table*}
\begin{table*}[!ht]
\centering
\caption{Results of Various CNNs on the  UCM Dataset with Greedy Search}
\label{UCM_greedy}
\resizebox{\textwidth}{!}{
\begin{tabular}{|c|c|c|c|c|c|c|c|c|}
\hline
CNN & BLEU-1 & BLEU-2 & BLEU-3 & BLEU-4 & METEOR & ROUGE-L & CIDEr & Cluster\\
\hline
ResNet & 0.8350 & 0.7686 & 0.7143 & 0.6658 & 0.4425 & 0.7822 & 3.4078 & Good\\
Wide ResNet & 0.8511 & 0.7856 & 0.7290 & 0.6777 & 0.4501 & 0.8071 & 3.5221 & Good\\
ResNext & 0.8210 & 0.7491 & 0.6905 & 0.6418 & 0.4332 & 0.7704 & 3.3276 & Medium\\
RegNet & 0.8079 & 0.7347 & 0.6747 & 0.6199 & 0.4326 & 0.7777 & 3.2392 & Medium\\
VGGNet & 0.8034 & 0.7286 & 0.6748 & 0.6321 & 0.4266 & 0.7585 & 3.2563 & Medium\\
DenseNet & 0.8199 & 0.7559 & 0.7032 & 0.6555 & 0.4565 & 0.7905 & 3.4760 & Good\\
AlexNet & 0.7785 & 0.6951 & 0.6327 & 0.5820 & 0.4024 & 0.7314 & 3.0124 & Bad\\
GoogleNet & 0.7855 & 0.7095 & 0.6516 & 0.5995 & 0.4161 & 0.7468 & 3.0661 & Bad\\
InceptionNet & 0.8338 & 0.7552 & 0.6919 & 0.6365 & 0.4477 & 0.8003 & 3.3445 & Medium\\
MobileNetV2 & 0.8203 & 0.7460 & 0.6884 & 0.6364 & 0.4291 & 0.7642 & 3.3664 & Medium\\
MobileNetV3 & 0.8088 & 0.7371 & 0.6825 & 0.6327 & 0.4208 & 0.7658 & 3.2499 & Medium\\
ConvNext & 0.8369 & 0.7712 & 0.7143 & 0.6612 & 0.4566 & 0.8119 & 3.4582 & Good\\
\hline
\end{tabular}}
\end{table*}
\begin{table*}[!ht]
\centering
\caption{Results of Various CNNs on the  UCM Dataset with Beam Search}
\label{UCM_beam}
\resizebox{\textwidth}{!}{
\begin{tabular}{|c|c|c|c|c|c|c|c|c|}
\hline
CNN & BLEU-1 & BLEU-2 & BLEU-3 & BLEU-4 & METEOR & ROUGE-L & CIDEr & Cluster\\
\hline
ResNet & 0.8434 & 0.7795 & 0.7259 & 0.6770 & 0.4523 & 0.7930 & 3.4061 & Good\\
Wide ResNet & 0.8459 & 0.7874 & 0.7339 & 0.6850 & 0.4591 & 0.8060 & 3.4676 & Good\\
ResNext & 0.8026 & 0.7312 & 0.6729 & 0.6229 & 0.4213 & 0.7564 & 3.1233 & Bad\\
RegNet & 0.8339 & 0.7662 & 0.7111 & 0.6604 & 0.4408 & 0.7809 & 3.2729 & Medium\\
VGGNet & 0.8054 & 0.7343 & 0.6775 & 0.6297 & 0.4389 & 0.7711 & 3.2628 & Bad\\
DenseNet & 0.8330 & 0.7741 & 0.7238 & 0.6767 & 0.4623 & 0.7983 & 3.4211 & Good\\
AlexNet & 0.8003 & 0.7240 & 0.6653 & 0.6158 & 0.4174 & 0.7401 & 3.0186 & Bad\\
GoogleNet & 0.7973 & 0.7285 & 0.6735 & 0.6243 & 0.4276 & 0.7573 & 3.0818 & Bad\\
InceptionNet & 0.8420 & 0.7663 & 0.7063 & 0.6527 & 0.4566 & 0.7988 & 3.3176 & Medium\\
MobileNetV2 & 0.8252 & 0.7603 & 0.7064 & 0.6563 & 0.4346 & 0.7655 & 3.3252 & Medium\\
MobileNetV3 & 0.8281 & 0.7609 & 0.7077 & 0.6590 & 0.4466 & 0.7829 & 3.2713 & Medium\\
ConvNext & 0.8442 & 0.7802 & 0.7319 & 0.6806 & 0.4602 & 0.8126 & 3.4953 & Good\\
\hline
\end{tabular}}
\end{table*}
\begin{table*}[!ht]
\centering
\caption{Results of Various CNNs on the  RSICD Dataset with Greedy Search}
\resizebox{\textwidth}{!}{
\label{RSICD_greedy}
\begin{tabular}{|c|c|c|c|c|c|c|c|c|}
\hline
CNN & BLEU-1 & BLEU-2 & BLEU-3 & BLEU-4 & METEOR & ROUGE-L & CIDEr & Cluster\\
\hline
ResNet & 0.6295 & 0.4607 & 0.3574 & 0.2878 & 0.2535 & 0.4671 & 0.8085 & Good\\
Wide ResNet & 0.6159 & 0.4451 & 0.3446 & 0.2788 & 0.2439 & 0.4551 & 0.7972 & Medium\\
ResNext & 0.6351 & 0.4579 & 0.3507 & 0.2811 & 0.2550 & 0.4703 & 0.8061 & Good\\
RegNet & 0.6231 & 0.4478 & 0.3439 & 0.2746 & 0.2462 & 0.4617 & 0.7971 & Good\\
VGGNet & 0.6111 & 0.4308 & 0.3252 & 0.2576 & 0.2390 & 0.4434 & 0.7321 & Bad\\
DenseNet & 0.6142 & 0.4376 & 0.3310 & 0.2603 & 0.2506 & 0.4552 & 0.7614 & Medium\\
AlexNet & 0.6046 & 0.4273 & 0.3227 & 0.2555 & 0.2345 & 0.4376 & 0.7092 & Bad\\
GoogleNet & 0.6213 & 0.4460 & 0.3417 & 0.2727 & 0.2445 & 0.4586 & 0.7787 & Medium\\
InceptionNet & 0.6146 & 0.4409 & 0.3370 & 0.2692 & 0.2443 & 0.4599 & 0.7839 & Medium\\
MobileNetV2 & 0.6174 & 0.4460 & 0.3416 & 0.2729 & 0.2471 & 0.4551 & 0.7618 & Medium\\
MobileNetV3 & 0.6209 & 0.4438 & 0.3386 & 0.2709 & 0.2454 & 0.4572 & 0.7706 & Medium\\
ConvNext & 0.6431 & 0.4665 & 0.3602 & 0.3013 & 0.2560 & 0.4945 & 0.8415 & Good\\
\hline
\end{tabular}}
\end{table*}
\begin{table*}[!ht]
\centering
\caption{Results of Various CNNs on the  RSICD Dataset with Beam Search}
\resizebox{\textwidth}{!}{
\label{RSICD_beam}
\begin{tabular}{|c|c|c|c|c|c|c|c|c|}
\hline
CNN & BLEU-1 & BLEU-2 & BLEU-3 & BLEU-4 & METEOR & ROUGE-L & CIDEr & Cluster\\
\hline
ResNet & 0.6314 & 0.4661 & 0.3642 & 0.2949 & 0.2548 & 0.4711 & 0.8243 & Good\\
Wide ResNet & 0.6162 & 0.4442 & 0.3428 & 0.2767 & 0.2436 & 0.4516 & 0.8070 & Bad\\
ResNext & 0.6310 & 0.4514 & 0.3431 & 0.2715 & 0.2531 & 0.4616 & 0.7971 & Medium\\
RegNet & 0.6182 & 0.4455 & 0.3429 & 0.2739 & 0.2462 & 0.4578 & 0.8112 & Bad\\
VGGNet & 0.6173 & 0.4449 & 0.3410 & 0.2731 & 0.2459 & 0.4519 & 0.7752 & Bad\\
DenseNet & 0.6264 & 0.4505 & 0.3450 & 0.2736 & 0.2588 & 0.4663 & 0.8020 & Medium\\
AlexNet & 0.6176 & 0.4396 & 0.3369 & 0.2697 & 0.2414 & 0.4472 & 0.7470 & Bad\\
GoogleNet & 0.6185 & 0.4452 & 0.3423 & 0.2735 & 0.2448 & 0.4579 & 0.7771 & Bad\\
InceptionNet & 0.6279 & 0.4530 & 0.3476 & 0.2780 & 0.2497 & 0.4659 & 0.8132 & Good\\
MobileNetV2 & 0.6273 & 0.4542 & 0.3491 & 0.2798 & 0.2539 & 0.4648 & 0.7991 & Good\\
MobileNetV3 & 0.6220 & 0.4435 & 0.3378 & 0.2691 & 0.2445 & 0.4546 & 0.7724 & Bad\\
ConvNext & 0.6380 & 0.4855 & 0.3782 & 0.3168 & 0.2869 & 0.4923 & 0.8506 & Good\\
\hline
\end{tabular}}
\end{table*}
In our work, we employ an encoder-decoder-based transformer model, where the encoder extracts image features that undergo cross-attention with the decoder to generate meaningful captions. On the encoder side, the patch size is set to $49$, and both the convolutional layer output and the transformer-based encoder output have a dimensionality of $768$. The multi-head encoder transformer comprises $16$ heads and is repeated six times. For the decoder, the number of hidden layers and attention heads is also set to $12$. The vocabulary size and maximum sequence length are determined based on the training dataset captions.

Training is carried out from start to finish using the AdamW optimizer with a learning rate of $1\times 10^{-4}$ and a batch size of $64$. The model is trained for a maximum of $64$ epochs, with early stopping applied if the ROUGE-L score in the validation set does not improve for $10$ consecutive epochs. A linear learning rate scheduler with a $10\%$ warm-up phase is used to facilitate convergence, and gradient clipping with a maximum norm of $1.0$ is applied to stabilize training. The categorical cross-entropy loss is used as the objective function. We follow the standard train-validation-test splits provided in each dataset. For caption generation during inference, the beam search is used with five beams, a length penalty of $2.0$, and n-gram repetition avoidance set to three. Our experiments were carried out on a Linux-based Ubuntu $22.04$ system equipped with~\emph{NVIDIA RTX A6000 GPU} $(48GB)$ and $124GB$ RAM. During the preparation of this manuscript, Grammarly and ChatGPT were employed exclusively for language refinement purposes, such as rephrasing and correcting grammatical or syntactic problems. These tools were not used to generate or contribute any original technical content.
\subsection{Dataset Utilized}
To evaluate the performance of our model, we used three well-established RSIC datasets. The details of these datasets are provided below.
\begin{itemize}
    \item\textbf{SYDNEY:} The SYDNEY dataset~\cite{qu2016deep} comprises a total of $613$ images, with $497$ designated for training, $58$ for testing and $58$ for validation. It originates from the Sydney dataset~\cite{zhang2014saliency} and has been meticulously refined to include seven distinct categories:~\emph{airport, industrial, meadow, ocean, residential, river,} and~\emph{runway}, achieved through careful selection and cropping.
    \item\textbf{UCM:} The UCM dataset~\cite{qu2016deep} consists of $2100$ images, with $1680$ allocated for training, $210$ for testing, and $210$ for validation. It is an adapted version of the~\emph{UC Merced Land Use} dataset~\cite{yang2010bag}, which contains $21$ land use categories, each comprising 100 images. These categories include~\emph{agriculture, airport, baseball diamond, beach, buildings, chaparral, denseresidential, forest, freeway, golfcourse, harbour, intersection, mediumresidential, mobilehomepark, overpass, parking, river, runway, sparseresidential, storagetanks,} and~\emph{tenniscourt}.
    \item\textbf{RSICD:} The RSICD dataset~\cite{lu2017exploring} contains an extensive collection of $10921$ images, with $8,034$ reserved for training, $1093$ for testing, and $1094$ for validation. Powered by multiple platforms, including Google Earth~\cite{xia2017aid}, Baidu Map, MapABC, and Tianditu, this dataset covers 31 distinct categories, such as~\emph{airport, bareland, baseballfield, beach, bridge, center, church, commercial, denseresidential, desert, farmland, forest, industrial, meadow, mediumresidential, mountain, park, parking, playground, pond, port, railwaystation, resort, river, school, sparseresidential, square, stadium, storagetanks,} and~\emph{viaduct}.
\end{itemize}
For this study, we used the corrected versions of these datasets~\cite{das2024textgcn}, which rectify common issues, including spelling inconsistencies, grammatical inaccuracies, and variations in English dialects. The train-validation-test split was maintained as originally defined in these datasets.
\subsection{Performance Metrics}  
The evaluation metrics used in our model are detailed below. These metrics are commonly used in RSIC~\cite{lu2017exploring,qu2016deep,hoxha2021novel}.
\begin{itemize}  
    \item\textbf{BLEU:} The Bilingual Evaluation Understudy (BLEU)~\cite{papineni-etal-2002-bleu} is a precision-based metric that assesses the similarity between a generated caption and reference captions by analyzing n-gram overlap. Calculate the geometric mean of the precisions in n gram while applying a brevity penalty to discourage excessively short outputs. BLEU is widely adopted in tasks such as machine translation and image captioning. In this study, we employ BLEU-1 to BLEU-4. 
    \item\textbf{METEOR:} The Metric for the Evaluation of Explicit Ordering Translations (METEOR)~\cite{lavie-agarwal-2007-meteor} evaluates generated captions considering factors such as stemming, synonym matching, and word order. Unlike BLEU, which is precision focused, METEOR integrates both precision and recall, assigning a higher weight to recall for improved accuracy.  
    \item\textbf{ROUGE:} The Recall-Oriented Understudy for Gisting Evaluation (ROUGE)~\cite{lin-2004-ROUGE} is a recall-based metric that measures n-gram overlap between generated and reference captions. In this work, we utilize ROUGE-L, which leverages the longest common subsequence (LCS) to assess text similarity.  
    \item\textbf{CIDEr:} The Consensus-based Image Description Evaluation (CIDEr)~\cite{vedantam2015cider} quantifies the quality of generated image captions by comparing them to a collection of human-written references. CIDEr evaluates how closely a generated caption aligns with the consensus of human descriptions, aiming to capture the overall semantic agreement between the generated text and human perception.
\end{itemize} 
\section{Results}
\label{sec_experimental}
\subsection{Numerical Evaluation of Different CNNs}

To enable a comparative analysis of CNN encoders based on their captioning effectiveness, we organized them into qualitative performance tiers. For this purpose, we classified CNN encoders into groups:~\emph{Good},~\emph{Medium}, and~\emph{Bad} using K-means clustering with $k=3$, based on seven standard captioning metrics: BLEU-1, BLEU-2, BLEU-3, BLEU-4, ROUGE-L, METEOR, and CIDEr. All metrics were treated equally to ensure that the clustering reflects captioning performance alone.

Since CIDEr scores typically range from $[0, 5]$ in standard benchmarks, we normalize them by dividing them by five to bring them on a scale comparable to the other metrics. This normalization ensured that all metrics contributed fairly during the clustering.

We then calculated the geometric mean of each cluster centroid to evaluate the overall quality of the group. The cluster with the highest geometric mean was designated as the~\emph{Good} cluster. To ensure robustness, we impose the restriction that the~\emph{Good} cluster must contain at least four encoders. If this condition was not satisfied, encoders from the other clusters closest to the~\emph{Good} centroid (based on the Euclidean distance) were reassigned accordingly. The remaining encoders were then reorganized into the~\emph{Medium} and~\emph{Bad} groups using a second K-means pass with $k=2$.

The performance of twelve encoder-decoder models across three datasets is presented in~\Cref{SYDNEY_greedy,SYDNEY_beam,UCM_greedy,UCM_beam,RSICD_greedy,RSICD_beam} using the seven evaluation metrics. It is observed that ConvNext and ResNet consistently belong to the~\emph{Good} group. In contrast, AlexNet performs poorly, remaining in the~\emph{Bad} cluster for all datasets except the SYDNEY dataset with greedy search, where it is assigned to the~\emph{Medium} cluster. GoogleNet and MobileNetV3 never appear in the~\emph{Good} cluster. RegNet, VGGNet, and MobileNetV2 appear once, while Wide ResNet, ResNext, and DenseNet appear twice. InceptionNet appears in the~\emph{Good} cluster three times. Furthermore, it can be concluded that the difference in performance between beam search and greedy decoding is not significant across all CNNs and datasets.

For each scenario, CNNs grouped under the~\emph{Good} group will undergo further evaluation through a human study, evaluating the generated captions from a subjective perspective (see~\Cref{sec_subjective}). The combined results of the numerical and human evaluations will provide insight into which CNN performs best across all conditions.

\subsection{Effectiveness of CNNs in Capturing the Core Meaning of Images}
\begin{table}[!ht]
\centering
\caption{Classification Performance Across Three Datasets (\%): \textcolor{redbox}{Red} denotes the first place,~\textcolor{bluebox}{Blue} denotes the second place, and~\textcolor{greenbox}{Green} denotes the third place}
\label{classification}
\resizebox{0.475\textwidth}{!}{
\begin{tabular}{|c|c|c|c|c|c|}
\hline
CNN & SYDNEY & UCM & RSICD & MACRO & MICRO \\
\hline
ResNet &~\textcolor{green}{91.38} & 91.43 &~\textcolor{green}{93.69} & 92.17 &~\textcolor{blue}{93.24} \\
Wide ResNet &~\textcolor{blue}{94.83} &~\textcolor{blue}{92.86} & 92.96 &~\textcolor{green}{93.55} & 93.02 \\
ResNext & 89.66 & 90.00 & 93.60 & 91.09 & 92.87 \\
RegNet & 89.66 & 90.00 &~\textcolor{blue}{93.96} & 91.21 &~\textcolor{green}{93.17} \\
VGGNet &~\textcolor{red}{96.55} & 86.19 & 89.75 & 90.83 & 89.49 \\
DenseNet &~\textcolor{red}{96.55} &~\textcolor{green}{91.90} & 93.32 &~\textcolor{blue}{93.92} &~\textcolor{blue}{93.24} \\
AlexNet &~\textcolor{blue}{94.83} & 89.52 & 85.45 & 89.93 & 86.48 \\
GoogleNet &~\textcolor{red}{96.55} & 90.95 & 91.22 & 92.91 & 91.40 \\
InceptionNet &~\textcolor{red}{96.55} & 87.14 & 92.22 & 91.97 & 91.62 \\
MobileNetV2 & 89.66 & 89.05 & 91.31 & 90.01 & 90.89 \\
MobileNetV3 &~\textcolor{green}{91.38} & 90.00 & 92.59 & 91.32 & 92.14 \\
ConvNext &~\textcolor{blue}{94.83} &~\textcolor{red}{93.81} &~\textcolor{red}{94.88} &~\textcolor{red}{94.51} &~\textcolor{red}{94.71} \\
\hline
\end{tabular}
}
\end{table}
The encoder plays a crucial role in RSIC. Its output significantly assists the decoder in generating accurate captions. The efficiency of an encoder depends on how well it interprets an image. A deep understanding of the core meaning is essential for an encoder, as remote sensing images exhibit intricate multi-scale structures and complex backgrounds. Proper interpretation enables the extraction of informative multiscale features, which are vital to generate precise and meaningful descriptions.  

To evaluate encoder performance, we performed a classification test on all encoders. The classification labels correspond to the image categories in each dataset, where SYDNEY has 7 classes, UCM has 21, and RSICD has 31. The classification model is defined by first taking the encoder's output in a 3D format (batch size, patch size, feature size) and then applying a weighted average to each patch to obtain a 2D representation (batch size, feature size). A linear layer is then added to map the features to the respective number of classes.  

\Cref{classification} presents the classification results for all CNNs across the three datasets, considering both micro and macro averages. For macro averaging, ConvNext ranks first, followed by DenseNet in second and Wide ResNet in third. In micro averaging, ConvNext retains the top position, while DenseNet and ResNet share second place, with RegNet in third. However, rankings differ across datasets. In the SYDNEY dataset, VGGNet, DenseNet, GoogleNet, and InceptionNet secure the first position, while Wide ResNet, AlexNet, and ConvNext rank second, and ResNet together with MobileNetV3 share the third position. The similarity in results among multiple CNNs in the Sydney dataset can be attributed to its small size and class imbalance. With only 58 test samples available, the model has limited variation for learning, resulting in uniform predictions across many CNNs~\cite{lu2017exploring}. This minimizes the likelihood of significant performance differences between them. In the UCM dataset, ConvNext holds the first position, Wide ResNet comes second, and DenseNet secures third. For the RSICD dataset, ConvNext ranks highest, followed by ResNet in second place and RegNet in third. These results highlight the variation in CNN performance across datasets, with ConvNext consistently demonstrating superior effectiveness.    
\subsection{Effect of Multi-Head Transformer After the CNN}
\begin{table*}[!ht]
\centering
\caption{Performace of Ablation Studies on the SYDNEY Dataset}
\label{SYDNEY_abl}
\resizebox{\textwidth}{!}{
\begin{tabular}{|c|c|c|c|c|c|c|c|}
\hline
Model & BLEU-1 & BLEU-2 & BLEU-3 & BLEU-4 & METEOR & ROUGE-L & CIDEr \\
\hline
R-WOT-G & 0.7195 & 0.6123 & 0.5286 & 0.4565 & 0.3566 & 0.6404 & 1.9592 \\
R-SHT-G & 0.7270 & 0.6331 & 0.5418 & 0.4864 & 0.3545 & 0.6671 & 2.0706 \\
R-MHT-G & 0.7480 & 0.6484 & 0.5719 & 0.5039 & 0.3630 & 0.6723 & 2.0973 \\
\hdashline
C-WOT-G & 0.7520 & 0.6648 & 0.5949 & 0.5364 & 0.3755 & 0.6945 & 2.3312 \\
C-SHT-G & 0.7455 & 0.6571 & 0.5818 & 0.5152 & 0.3856 & 0.7129 & 2.3365 \\
C-MHT-G &~\textbf{0.7997} &~\textbf{0.6844} & \textbf{0.6325} & \textbf{0.5694} & \textbf{0.4073} & \textbf{0.7349} & \textbf{2.4945} \\
\hdashline
R-WOT-B & 0.7031 & 0.6004 & 0.5238 & 0.4625 & 0.3476 & 0.6437 & 1.9377 \\
R-SHT-B & 0.7150 & 0.6154 & 0.5393 & 0.4735 & 0.3620 & 0.6573 & 2.0411 \\
R-MHT-B & 0.7578 & 0.6663 & 0.5906 & 0.5228 & 0.3729 & 0.6883 & 2.1498 \\
\hdashline
C-WOT-B & 0.7250 & 0.6322 & 0.5567 & 0.4901 & 0.3639 & 0.6651 & 2.1380 \\
C-SHT-B & 0.7348 & 0.6475 & 0.5702 & 0.4989 & 0.3684 & 0.6713 & 2.1677 \\
C-MHT-B & 0.7762 & 0.6742 & 0.6255 & 0.5513 & 0.3955 & 0.7281 & 2.4176 \\
\hline
\end{tabular}}
\end{table*}
\begin{table*}[!ht]
\centering
\caption{Performace of Ablation Studies on the UCM Dataset}
\label{UCM_abl}
\resizebox{\textwidth}{!}{
\begin{tabular}{|c|c|c|c|c|c|c|c|}
\hline
Model & BLEU-1 & BLEU-2 & BLEU-3 & BLEU-4 & METEOR & ROUGE-L & CIDEr \\
\hline
R-WOT-G & 0.8105 & 0.7430 & 0.6897 & 0.6406 & 0.4373 & 0.7604 & 3.3090 \\
R-SHT-G & 0.8269 & 0.7518 & 0.6961 & 0.6471 & 0.4331 & 0.7674 & 3.3512 \\
R-MHT-G & 0.8350 & 0.7686 & 0.7143 & 0.6658 & 0.4425 & 0.7822 & 3.4078 \\
\hdashline
C-WOT-G & 0.7805 & 0.7014 & 0.6385 & 0.5829 & 0.4124 & 0.7354 & 3.0045 \\
C-SHT-G & 0.8197 & 0.7468 & 0.6873 & 0.6360 & 0.4448 & 0.7871 & 3.3232 \\
C-MHT-G & 0.8369 & 0.7712 & 0.7143 & 0.6612 & 0.4566 & 0.8119 & 3.4582 \\
\hdashline
R-WOT-B & 0.8283 & 0.7654 & 0.7124 & 0.6621 & 0.4363 & 0.7797 & 3.2854 \\
R-SHT-B & 0.8244 & 0.7559 & 0.7001 & 0.6502 & 0.4524 & 0.7867 & 3.3968 \\
R-MHT-B & 0.8394 & 0.7795 & 0.7259 & 0.6770 & 0.4523 & 0.7930 & 3.4061 \\
\hdashline
C-WOT-B & 0.8179 & 0.7569 & 0.6955 & 0.6178 & 0.4253 & 0.7545 & 3.0367 \\
C-SHT-B & 0.8339 & 0.7743 & 0.7065 & 0.6453 & 0.4554 & 0.8053 & 3.4370 \\
C-MHT-B & \textbf{0.8442} & \textbf{0.7802} & \textbf{0.7319} &~\textbf{0.6806} & \textbf{0.4602} & \textbf{0.8126} & \textbf{3.4953} \\
\hline
\end{tabular}}
\end{table*}
\begin{table*}[!ht]
\centering
\caption{Performace of Ablation Studies on the RSICD Dataset}
\label{RSICD_abl}
\resizebox{\textwidth}{!}{
\begin{tabular}{|c|c|c|c|c|c|c|c|}
\hline
Model & BLEU-1 & BLEU-2 & BLEU-3 & BLEU-4 & METEOR & ROUGE-L & CIDEr \\
\hline
R-WOT-G & 0.6127 & 0.4368 & 0.3312 & 0.2627 & 0.2325 & 0.4261 & 0.7921 \\
R-SHT-G & 0.6199 & 0.4413 & 0.3420 & 0.2772 & 0.2311 & 0.4413 & 0.7815 \\
R-MHT-G & 0.6295 & 0.4607 & 0.3574 & 0.2878 & 0.2535 & 0.4671 & 0.8085 \\
\hdashline
C-WOT-G & 0.6176 & 0.4540 & 0.3477 & 0.2866 & 0.2503 & 0.4798 & 0.8197 \\
C-SHT-G & 0.6280 & 0.4607 & 0.3505 & 0.2892 & 0.2614 & 0.4862 & 0.8279 \\
C-MHT-G & \textbf{0.6431} & 0.4665 & 0.3602 & 0.3013 & 0.2560 & 0.4945 & 0.8415 \\
\hdashline
R-WOT-B & 0.6121 & 0.4409 & 0.3385 & 0.2708 & 0.2579 & 0.4609 & 0.7594 \\
R-SHT-B & 0.6171 & 0.4594 & 0.3318 & 0.2737 & 0.2422 & 0.4556 & 0.8054 \\
R-MHT-B & 0.6314 & 0.4661 & 0.3642 & 0.2949 & 0.2548 & 0.4711 & 0.8243 \\
\hdashline
C-WOT-B & 0.6294 & 0.4684 & 0.3612 & 0.2994 & 0.2668 & 0.4801 & 0.8336 \\
C-SHT-B & 0.6231 & 0.4648 & 0.3592 & 0.3084 & 0.2784 & 0.4859 & 0.8257 \\
C-MHT-B & 0.6380 & \textbf{0.4855} & \textbf{0.3782} & \textbf{0.3168} & \textbf{0.2869} & \textbf{0.4923} & \textbf{0.8506} \\
\hline
\end{tabular}}
\end{table*}
We conducted an ablation study to evaluate the impact of incorporating a multi-head transformer in our model. For this purpose, we performed experiments with three different encoder setups. In the first set-up, the encoder output was taken directly from the convolution layer. In the second setup, a single-head transformer was used in the encoder. Finally, in the third architecture, a multi-head transformer was employed.  

The numerical comparisons of these ablation studies are presented in~\Cref{SYDNEY_abl,UCM_abl,RSICD_abl}.The notation used in these tables follows the convention:~\emph{Encoder-Orientation-Search}.
\begin{itemize}
    \item\emph{Encoder}: Indicates the CNN used in the encoder. Here,~\emph{C} refers to ConvNext and~\emph{R} refers to ResNet. The reason behind selecting these two CNNs is that they are included in~\emph{Good} cluster for all the cases in the numerical evaluation.
    \item\emph{Orientation}: Indicates the orientation of the encoder used in the model. In these tables,~\emph{WOT} denotes that the output of the encoder is obtained from the convolution layer (\Cref{architecture}),~\emph{SHT} denotes that the encoder transformer has a single head, and~\emph{MHT} denotes that the encoder transformer is multi-headed.
    \item\emph{Search}: Indicates the search technique used to generate the caption. Here,~\emph{G} represents greedy search, and~\emph{B} represents beam search.
\end{itemize}
The results clearly demonstrate that integrating a transformer after the CNN enhances performance compared to using only a CNN. Specifically, the~\emph{Single-Head Transformer (SHT)} setup outperforms the~\emph{Without Transformer (WOT)} approach, highlighting the benefits of self-attention in capturing long-range dependencies that standard CNNs alone struggle with. While CNNs are effective at extracting local features, they lack the ability to model global context efficiently. By introducing a self-attention mechanism, SHT enables the model to focus on different regions of the image dynamically, leading to improved feature representation and better caption generation.  

Furthermore,~\emph{Multi-Head Transformer (MHT)} consistently surpasses both the WOT and SHT configurations. Unlike SHT, which uses a single attention head, MHT employs multiple attention heads that can attend to different parts of the image simultaneously. This allows the model to capture a richer set of dependencies, improving its ability to generate more contextually accurate and semantically meaningful captions. The parallel attention mechanism in MHT enhances the diversity of features, ensuring that local and global relationships within the image are effectively learned. These improvements demonstrate that while self-attention alone improves performance over a pure CNN encoder, the use of multiple attention heads further strengthens the model’s ability to process complex image structures, leading to better captioning results.
\subsection{Comparison of Various Encoder–decoder Combinations}
\begin{table}[!ht]
\centering
\caption{Performance of Various Encoder-decoder Combinations on the SYDNEY Dataset with Greedy Search}
\label{SYDNEY_pair_greedy}
\resizebox{\textwidth}{!}{
\begin{tabular}{|c|c|c|c|c|c|c|c|c|}
\hline
\textbf{Encoder} & \textbf{Decoder} & \textbf{BLEU-1} & \textbf{BLEU-2} & \textbf{BLEU-3} & \textbf{BLEU-4} & \textbf{METEOR} & \textbf{ROUGE-L} & \textbf{CIDEr} \\
\hline
\multirow{3}{*}{ConvNext}
 & Roberta & 0.7717 & 0.6663 & 0.6298 & 0.5631 & 0.4036 & \textbf{0.7393} & \textbf{2.5433} \\
 & BART & 0.7556 & 0.6492 & 0.6057 & 0.5438 & 0.3922 & 0.7253 & 2.4152 \\
 & GPT2 & \textbf{0.7997} & \textbf{0.6844} & \textbf{0.6325} & \textbf{0.5694} & 0.4073 & 0.7349 & 2.4945 \\
\hdashline
\multirow{3}{*}{ResNet}
 & Roberta & 0.7598 & 0.6503 & 0.6231 & 0.5594 & 0.3839 & 0.7198 & 2.3544 \\
 & BART & 0.7440 & 0.6350 & 0.6189 & 0.5514 & 0.3674 & 0.7079 & 2.3093 \\
 & GPT2 & 0.7670 & 0.6784 & 0.6281 & 0.5639 & 0.3896 & 0.7223 & 2.3873 \\
\hdashline
\multirow{3}{*}{Swin}
 & Roberta & 0.7383 & 0.6455 & 0.5745 & 0.5166 & 0.3691 & 0.6681 & 2.3116 \\
 & BART & 0.7443 & 0.6666 & 0.6047 & 0.5497 & 0.3720 & 0.6807 & 2.2430 \\
 & GPT2 & 0.7551 & 0.6676 & 0.6104 & 0.5541 & 0.3700 & 0.6963 & 2.3433 \\
\hdashline
\multirow{3}{*}{ViT} 
 & Roberta & 0.7304 & 0.6353 & 0.5833 & 0.5217 & 0.3849 & 0.6724 & 2.0215 \\
 & BART & 0.7414 & 0.6591 & 0.6189 & 0.5574 & 0.4011 & 0.7146 & 2.1846 \\
 & GPT2 & 0.7493 & 0.6687 & 0.6251 & 0.5629 & \textbf{0.4125} & 0.7274 & 2.2601 \\
\hline
\end{tabular}}
\end{table}
\begin{table}[!ht]
\centering
\caption{Performance of Various Encoder--decoder Combinations on the SYDNEY Dataset with Beam Search}
\label{SYDNEY_pair_beam}
\resizebox{\textwidth}{!}{
\begin{tabular}{|c|c|c|c|c|c|c|c|c|}
\hline
\textbf{Encoder} & \textbf{Decoder} & \textbf{BLEU-1} & \textbf{BLEU-2} & \textbf{BLEU-3} & \textbf{BLEU-4} & \textbf{METEOR} & \textbf{ROUGE-L} & \textbf{CIDEr} \\
\hline
\multirow{3}{*}{ConvNext} & Roberta & 0.7433 & 0.6537 & 0.5742 & 0.5061 & 0.3741 & 0.6806 & 2.2918 \\
 & BART & 0.7642 & 0.6658 & 0.5926 & 0.5176 & 0.4063 & 0.7357 & 2.3668 \\
 & GPT2 & \textbf{0.7762} & \textbf{0.6742} & \textbf{0.6255} & \textbf{0.5513} & \textbf{0.3955} & \textbf{0.7281} & \textbf{2.4176} \\
\hdashline
\multirow{3}{*}{ResNet} & Roberta & 0.7229 & 0.6302 & 0.5873 & 0.4846 & 0.3588 & 0.6545 & 2.0592 \\
 & BART & 0.7375 & 0.6444 & 0.5607 & 0.5089 & 0.3669 & 0.6673 & 2.0915 \\
 & GPT2 & 0.7578 & 0.6663 & 0.5906 & 0.5228 & 0.3729 & 0.6883 & 2.1498 \\
\hdashline
\multirow{3}{*}{Swin} & Roberta & 0.7052 & 0.6134 & 0.5465 & 0.4908 & 0.3630 & 0.6466 & 1.8968 \\
 & BART & 0.7132 & 0.6359 & 0.5756 & 0.5265 & 0.3731 & 0.6765 & 2.2107 \\
 & GPT2 & 0.7240 & 0.6414 & 0.5740 & 0.5174 & 0.3797 & 0.6823 & 2.2496 \\
\hdashline
\multirow{3}{*}{ViT} & Roberta & 0.7175 & 0.6035 & 0.5638 & 0.4859 & 0.3700 & 0.6627 & 2.0164 \\
 & BART & 0.7303 & 0.6194 & 0.5692 & 0.4915 & 0.3773 & 0.6821 & 2.1348 \\
 & GPT2 & 0.7386 & 0.6237 & 0.5781 & 0.5092 & 0.3747 & 0.6795 & 2.1799 \\
\hline
\end{tabular}}
\end{table}
\begin{table}[!ht]
\centering
\caption{Performance of Various Encoder-decoder Combinations on the UCM Dataset with Greedy Search}
\label{UCM_pair_greedy}
\resizebox{\textwidth}{!}{
\begin{tabular}{|c|c|c|c|c|c|c|c|c|}
\hline
\textbf{Encoder} & \textbf{Decoder} & \textbf{BLEU-1} & \textbf{BLEU-2} & \textbf{BLEU-3} & \textbf{BLEU-4} & \textbf{METEOR} & \textbf{ROUGE-L} & \textbf{CIDEr} \\
\hline
\multirow{3}{*}{ConvNext}
 & Roberta & 0.8109 & 0.7560 & 0.7043 & 0.6587 & 0.4315 & 0.7858 & 3.3585 \\
 & BART & 0.8249 & 0.7619 & 0.7092 & 0.6596 & 0.4492 & 0.7943 & 3.4335 \\
 & GPT2 & \textbf{0.8369} & \textbf{0.7712} & 0.7143 & 0.6612 & \textbf{0.4566} & \textbf{0.8119} & \textbf{3.4582} \\
\hdashline
\multirow{3}{*}{ResNet}
 & Roberta & 0.8205 & 0.7507 & 0.6967 & 0.6498 & 0.4304 & 0.7736 & 3.3684 \\
 & BART & 0.8330 & 0.7692 & \textbf{0.7152} & 0.6648 & 0.4379 & 0.7868 & 3.3022 \\
 & GPT2 & 0.8350 & 0.7686 & 0.7143 & \textbf{0.6658} & 0.4425 & 0.7822 & 3.4078 \\
\hdashline
\multirow{3}{*}{Swin}
 & Roberta & 0.7977 & 0.7247 & 0.6658 & 0.6146 & 0.4335 & 0.7679 & 3.2830 \\
 & BART & 0.8125 & 0.7345 & 0.6735 & 0.6218 & 0.4418 & 0.7737 & 3.3439 \\
 & GPT2 & 0.8245 & 0.7563 & 0.7020 & 0.6523 & 0.4367 & 0.7861 & 3.3869 \\
\hdashline
\multirow{3}{*}{ViT}
 & Roberta & 0.8009 & 0.7460 & 0.6943 & 0.6487 & 0.4215 & 0.7758 & 3.1160 \\
 & BART & 0.8129 & 0.7491 & 0.6894 & 0.6349 & 0.4061 & 0.7641 & 3.0014 \\
 & GPT2 & 0.8218 & 0.7568 & 0.7031 & 0.6517 & 0.4301 & 0.7832 & 3.2982 \\
\hline
\end{tabular}}
\end{table}
\begin{table}[!ht]
\centering
\caption{Performance of Various Encoder-decoder Combinations on the UCM Dataset with Beam Search}
\label{UCM_pair_beam}
\resizebox{\textwidth}{!}{
\begin{tabular}{|c|c|c|c|c|c|c|c|c|}
\hline
\textbf{Encoder} & \textbf{Decoder} & \textbf{BLEU-1} & \textbf{BLEU-2} & \textbf{BLEU-3} & \textbf{BLEU-4} & \textbf{METEOR} & \textbf{ROUGE-L} & \textbf{CIDEr} \\
\hline
\multirow{3}{*}{ConvNext}
 & Roberta & 0.8012 & 0.7516 & 0.6991 & 0.6451 & 0.4349 & 0.7790 & 3.1578 \\
 & BART & 0.8099 & 0.7513 & 0.7033 & 0.6549 & 0.4394 & 0.7711 & 3.0948 \\
 & GPT2 & \textbf{0.8223} & \textbf{0.7589} & \textbf{0.7061} & \textbf{0.6581} & 0.4438 & \textbf{0.7810} & 3.3226 \\
\hdashline
\multirow{3}{*}{ResNet}
 & Roberta & 0.7946 & 0.7323 & 0.6713 & 0.6278 & 0.4159 & 0.7549 & 3.2378 \\
 & BART & 0.8011 & 0.7401 & 0.6872 & 0.6493 & 0.4448 & 0.7771 & 3.2759 \\
 & GPT2 & 0.8115 & 0.7428 & 0.6883 & 0.6418 & 0.4373 & 0.7750 & 3.2871 \\
\hdashline
\multirow{3}{*}{Swin}
 & Roberta & 0.7752 & 0.7103 & 0.6544 & 0.6008 & 0.4273 & 0.7563 & 3.1882 \\
 & BART & 0.7930 & 0.7322 & 0.6824 & 0.6360 & \textbf{0.4468} & 0.7763 & 3.2705 \\
 & GPT2 & 0.7977 & 0.7266 & 0.6698 & 0.6227 & 0.4504 & 0.7724 & \textbf{3.3430} \\
\hdashline
\multirow{3}{*}{ViT}
 & Roberta & 0.7940 & 0.7271 & 0.6718 & 0.6195 & 0.4153 & 0.7659 & 3.0635 \\
 & BART & 0.7942 & 0.7351 & 0.6848 & 0.6378 & 0.4212 & 0.7575 & 3.1465 \\
 & GPT2 & 0.8033 & 0.7405 & 0.6873 & 0.6357 & 0.4368 & 0.7726 & 3.1967 \\
\hline
\end{tabular}}
\end{table}
\begin{table}[!ht]
\centering
\caption{Performance of Various Encoder-decoder Combinations on the RSICD Dataset with greedy Search}
\label{RSICD_pair_greedy}
\resizebox{\textwidth}{!}{
\begin{tabular}{|c|c|c|c|c|c|c|c|c|}
\hline
\textbf{Encoder} & \textbf{Decoder} & \textbf{BLEU-1} & \textbf{BLEU-2} & \textbf{BLEU-3} & \textbf{BLEU-4} & \textbf{METEOR} & \textbf{ROUGE-L} & \textbf{CIDEr} \\
\hline
\multirow{3}{*}{ConvNext}
 & Roberta & 0.6333 & 0.4578 & 0.3532 & 0.2916 & 0.2454 & 0.4716 & 0.8131 \\
 & BART & 0.6230 & 0.4464 & 0.3377 & 0.2635 & 0.2420 & 0.4576 & 0.8088 \\
 & GPT2 & \textbf{0.6431} & \textbf{0.4665} & \textbf{0.3602} & \textbf{0.3013} & 0.2560 & \textbf{0.4945} & \textbf{0.8415} \\
\hdashline
\multirow{3}{*}{ResNet}
 & Roberta & 0.6232 & 0.4501 & 0.3454 & 0.2753 & 0.2557 & 0.4639 & 0.7903 \\
 & BART & 0.6101 & 0.4349 & 0.3302 & 0.2595 & 0.2525 & 0.4529 & 0.7338 \\
 & GPT2 & 0.6295 & 0.4607 & 0.3574 & 0.2878 & 0.2535 & 0.4671 & 0.8085 \\
\hdashline
\multirow{3}{*}{Swin}
 & Roberta & 0.6160 & 0.4489 & 0.3459 & 0.2767 & 0.2428 & 0.4597 & 0.7535 \\
 & BART & 0.6028 & 0.4289 & 0.3219 & 0.2488 & 0.2305 & 0.4461 & 0.7181 \\
 & GPT2 & 0.6224 & 0.4431 & 0.3343 & 0.2610 & 0.2315 & 0.4491 & 0.7877 \\
\hdashline
\multirow{3}{*}{ViT}
 & Roberta & 0.6233 & 0.4536 & 0.3496 & 0.2802 & 0.2561 & 0.4700 & 0.7486 \\
 & BART & 0.6272 & 0.4517 & 0.3430 & 0.2714 & 0.2499 & 0.4594 & 0.7237 \\
 & GPT2 & 0.6332 & 0.4621 & 0.3573 & 0.2879 & \textbf{0.2616} & 0.4751 & 0.7567 \\
\hline
\end{tabular}}
\end{table}
\begin{table}[!ht]
\centering
\caption{Performance of Various Encoder-decoder Combinations on the RSICD Dataset with Beam Search}
\label{RSICD_pair_beam}
\resizebox{\textwidth}{!}{
\begin{tabular}{|c|c|c|c|c|c|c|c|c|}
\hline
\textbf{Encoder} & \textbf{Decoder} & \textbf{BLEU-1} & \textbf{BLEU-2} & \textbf{BLEU-3} & \textbf{BLEU-4} & \textbf{METEOR} & \textbf{ROUGE-L} & \textbf{CIDEr} \\
\hline
\multirow{3}{*}{ConvNext}
 & Roberta & 0.6245 & 0.4615 & 0.3589 & 0.2885 & 0.2606 & 0.4765 & 0.8165 \\
 & BART & 0.6182 & 0.4466 & 0.3428 & 0.2720 & 0.2578 & 0.4755 & 0.7834 \\
 & GPT2 & \textbf{0.6380} & \textbf{0.4855} & \textbf{0.3782} & \textbf{0.3168} & \textbf{0.2869} & \textbf{0.4923} & \textbf{0.8506} \\
\hdashline
\multirow{3}{*}{ResNet}
 & Roberta & 0.6234 & 0.4534 & 0.3505 & 0.2813 & 0.2418 & 0.4681 & 0.8191 \\
 & BART & 0.6169 & 0.4444 & 0.3411 & 0.2897 & 0.2478 & 0.4729 & 0.8063 \\
 & GPT2 & 0.6314 & 0.4661 & 0.3642 & 0.2949 & 0.2548 & 0.4711 & 0.8243 \\
\hdashline
\multirow{3}{*}{Swin}
 & Roberta & 0.6206 & 0.4575 & 0.3557 & 0.2873 & 0.2666 & 0.4780 & 0.7944 \\
 & BART & 0.6218 & 0.4471 & 0.3424 & 0.2723 & 0.2525 & 0.4627 & 0.7846 \\
 & GPT2 & 0.6244 & 0.4664 & 0.3571 & 0.2837 & 0.2702 & 0.4798 & 0.8019 \\
\hdashline
\multirow{3}{*}{ViT}
 & Roberta & 0.6149 & 0.4519 & 0.3450 & 0.2750 & 0.2587 & 0.4675 & 0.7818 \\
 & BART & 0.6190 & 0.4633 & 0.3475 & 0.2760 & 0.2709 & 0.4797 & 0.7768 \\
 & GPT2 & 0.6289 & 0.4747 & 0.3597 & 0.2914 & 0.2759 & 0.4882 & 0.8104 \\
\hline
\end{tabular}}
\end{table}
To further explore and justify the effectiveness of different components in our RSIC model, we conducted experiments using various encoder–decoder pairings. Specifically, we evaluated ConvNeXt and ResNet encoders, both of which consistently demonstrated strong performance in earlier experiments~\Cref{SYDNEY_greedy,SYDNEY_beam,UCM_greedy,UCM_beam,RSICD_greedy,RSICD_beam}, in addition to alternative vision transformers such as ViT~\cite{dosovitskiy2020image} and Swin~\cite{liu2021swin}. On the decoder side, in addition to GPT2~\cite{radford2019language} used in our baseline model, we tested RoBERTa~\cite{liu2019Roberta} and BART~\cite{lewis2019bart} to examine the influence of different text generation strategies.~\Cref{arch_sydney,arch_ucm,arch_rsicd} presents a comparative overview of key architectural attributes of the different decoders, including vocabulary size, maximum sequence length, FLOPs, and parameter count. Here,~\emph{Vocab Size} refers to the number of tokens in the decoder's vocabulary;\emph{Max Len} denotes the maximum sequence length supported by the decoder;~\emph{FLOPs} indicates the number of floating-point operations (in billions); and~\emph{\#PARAMS} represents the total number of decoder parameters (also in billions). All decoders use the same ByteLevelBPETokenizer~\cite{gage1994new,sennrich2015neural}, which standardizes the tokenization of input. As a result, the vocabulary size and the maximum sequence length remain identical across the decoders for a given dataset. Variations in FLOPs and parameter counts across datasets arise from decoder-specific fine-tuning performed separately for each dataset.

The comparative performance of these combinations is presented in~\Cref{SYDNEY_pair_greedy,SYDNEY_pair_beam,UCM_pair_greedy,UCM_pair_beam,RSICD_pair_greedy,RSICD_pair_beam}. Compared with earlier CNN-only results~\Cref{SYDNEY_greedy,SYDNEY_beam,UCM_greedy,UCM_beam,RSICD_greedy,RSICD_beam}, we observe that although ViT and Swin sometimes outperform CNN-based models in~\emph{medium} or~\emph{bad} clusters, carefully chosen CNNs such as ConvNext and ResNet still surpass them in most scenarios. This can be attributed to the fact that high-performing CNNs provide spatially rich and locally attentive features that align well with the characteristics of remote sensing images, which often contain strong local patterns, repetitive structures, and multi-scale objects. Vision transformers, in contrast, treat the image more uniformly and may overlook region-specific features that are crucial in aerial imagery due to the absence of an inductive bias for spatial locality.
\begin{table*}[!ht]
\begin{center}
\setlength{\tabcolsep}{0.3mm}
\tiny
\begin{minipage}{0.32\textwidth}
    \caption{Computational Complexities of Different Decoders on the SYDNEY Dataset}
    \label{arch_sydney}
    \resizebox{\textwidth}{!}{
    \begin{tabular}{|c|c|c|c|c|}
    \hline
    Decoder & Vocab Size & Max Len & FLOPs & \#PARAMS \\
    \hline
    Roberta & \multirow{3}{*}{480} & \multirow{3}{*}{40} & 6.8798 & 0.1144 \\
    BART    &                       &                     & 12.1298 & 0.2021 \\
    GPT2    &                       &                     & 6.8260 & 0.1138 \\
    \hline
    \end{tabular}}
\end{minipage}
\hfill
\begin{minipage}{0.32\textwidth}
    \caption{Computational Complexities of Different Decoders on the UCM Dataset}
    \label{arch_ucm}
    \resizebox{\textwidth}{!}{
    \begin{tabular}{|c|c|c|c|c|}
    \hline
    Decoder & Vocab Size & Max Len & FLOPs & \#PARAMS \\
    \hline
    Roberta & \multirow{3}{*}{745} & \multirow{3}{*}{44} & 7.5858 & 0.1146 \\
    BART    &                       &                     & 13.3667 & 0.2024 \\
    GPT2    &                       &                     & 7.5265 & 0.1140 \\
    \hline
    \end{tabular}}
\end{minipage}
\hfill
\begin{minipage}{0.32\textwidth}
    \caption{Computational Complexities of Different Decoders on the RSICD Dataset}
    \label{arch_rsicd}
    \resizebox{\textwidth}{!}{
    \begin{tabular}{|c|c|c|c|c|}
    \hline
    Decoder & Vocab Size & Max Len & FLOPs & \#PARAMS \\
    \hline
    Roberta & \multirow{3}{*}{5240} & \multirow{3}{*}{68} & 12.1936 & 0.1181 \\
    BART    &                        &                     & 21.1896 & 0.2063 \\
    GPT2    &                        &                     & 12.1013 & 0.1175 \\
    \hline
    \end{tabular}}
\end{minipage}
\end{center}
\end{table*}
Furthermore, our experiments show that GPT2 consistently outperforms other decoders (Roberta and BART) across all datasets and encoder setups. The autoregressive structure of GPT2 models caption generation as a sequential process, and its causal attention allows each token to be generated based on the full context of previous tokens. This helps maintain fluency and coherence across the caption. Its generation style suits the demands of remote sensing, where captions require accurate word order and semantic consistency. Although all decoders use cross-attention to incorporate visual information, GPT2 seems to utilize this visual grounding more effectively during generation. Its training objective encourages a tighter alignment between the preceding context and current prediction, enhancing its ability to integrate visual semantics at each step. These factors collectively contribute to the consistent superiority of GPT2 across encoder pairings.

In summary, we find that the combination of ConvNext as the encoder and GPT2 as the decoder delivers the most robust performance across all datasets and decoding strategies. This pairing consistently outperforms other combinations, highlighting the importance of jointly selecting encoder and decoder architectures for remote sensing captioning.
\subsection{Effect of Beam Size in Decoding Performance}
\begin{figure*}[!ht]
  \centering
  \subfloat[ConvNext-SYDNEY\label{ConvNext-SYDNEY}]{\includegraphics[width=0.3\textwidth,height=125px]{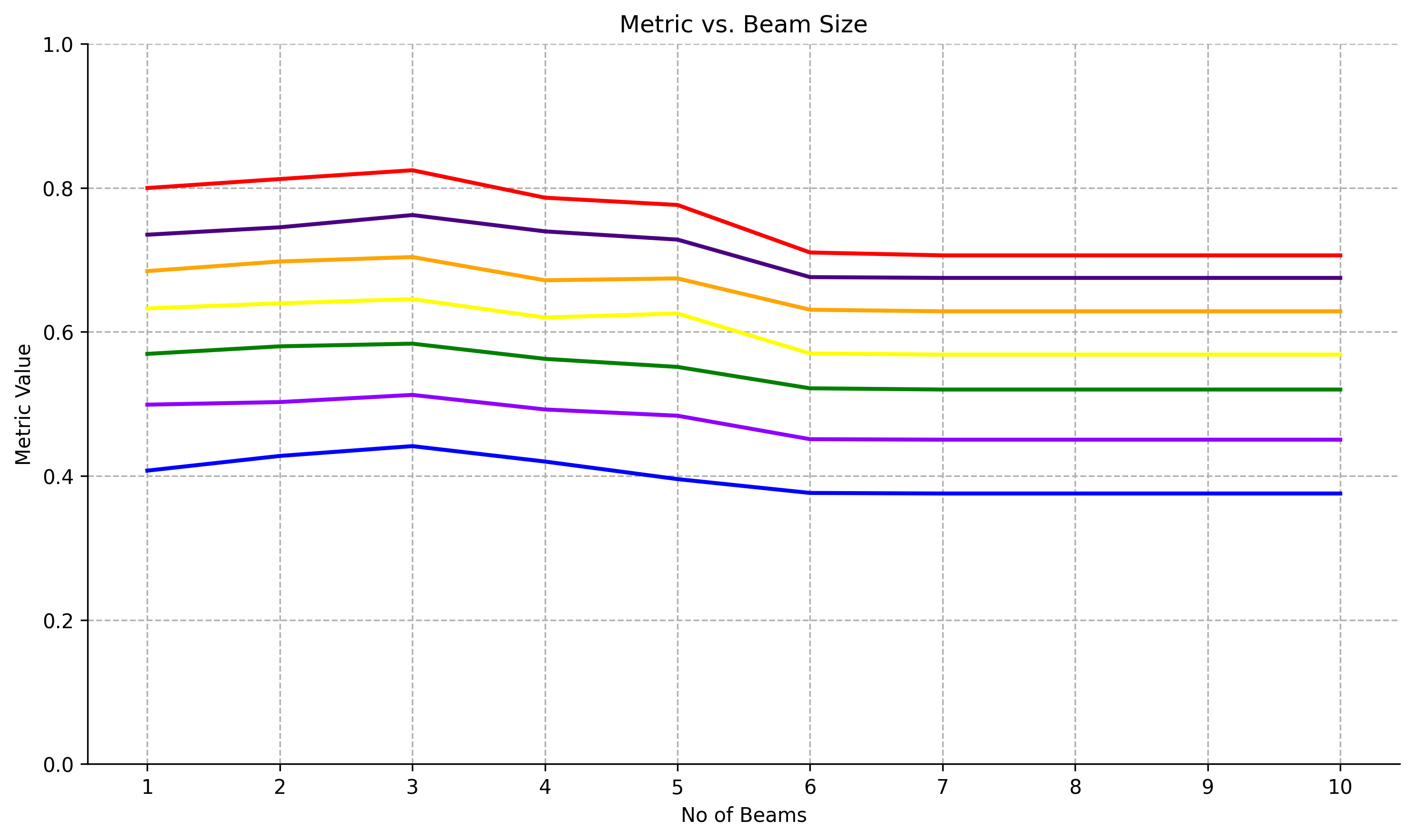}}\hfill
  \subfloat[ResNet-SYDNEY\label{ResNet-SYDNEY}]{\includegraphics[width=0.3\textwidth,height=125px]{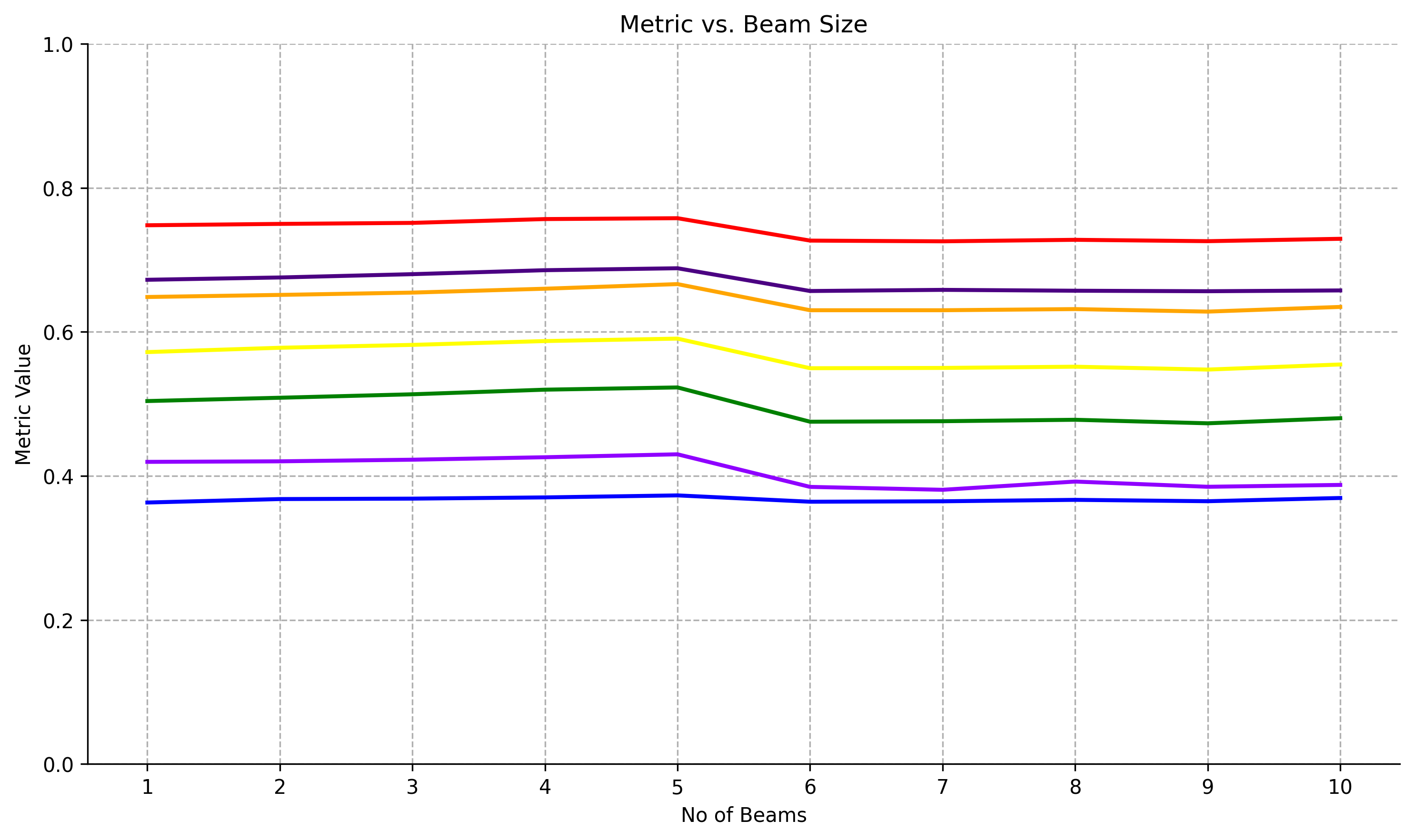}}\hfill
  \subfloat[ConvNext-UCM\label{ConvNext-UCM}]{\includegraphics[width=0.3\textwidth,height=125px]{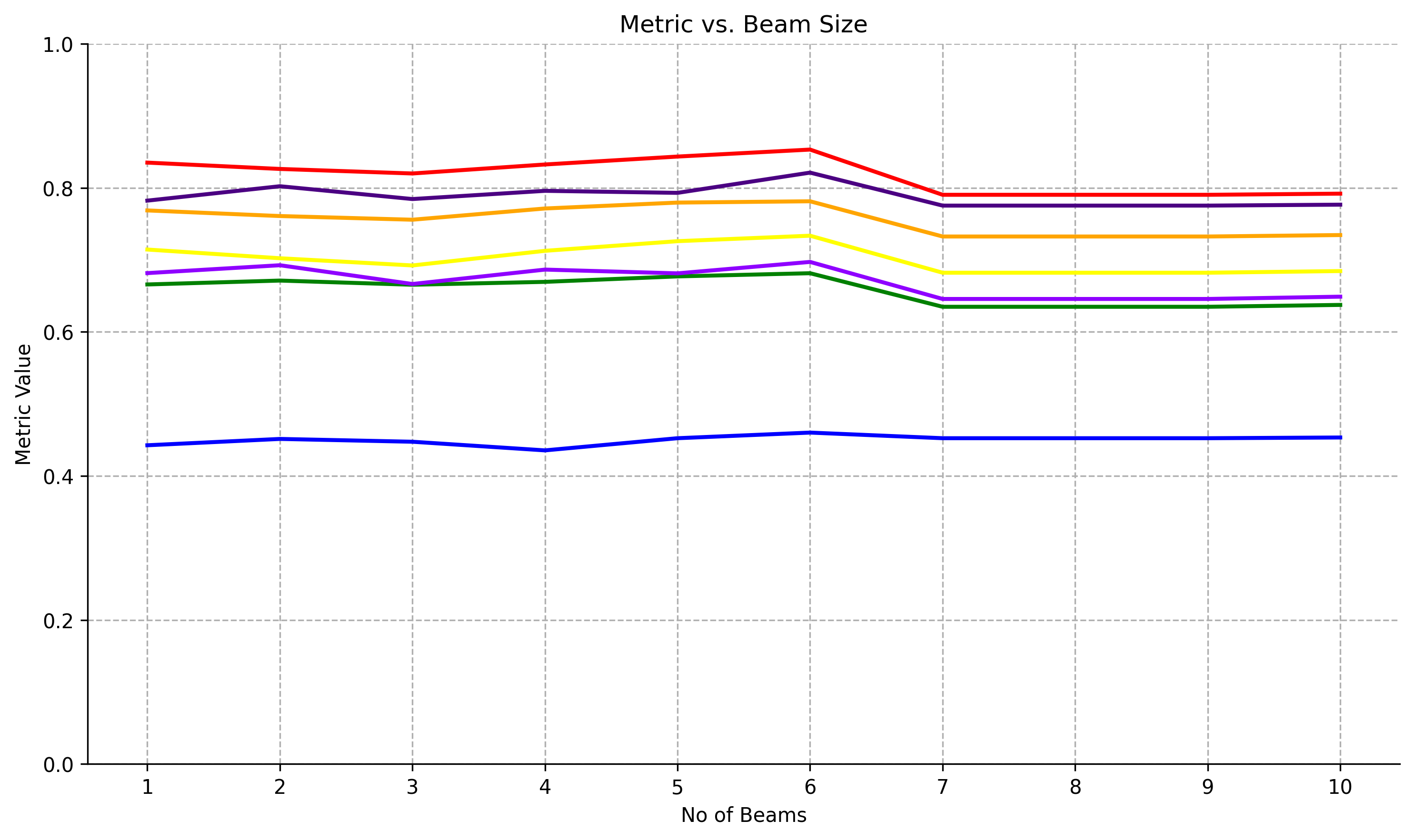}}\\[1ex]
  \subfloat[ResNet-UCM\label{ResNet-UCM}]{\includegraphics[width=0.3\textwidth,height=125px]{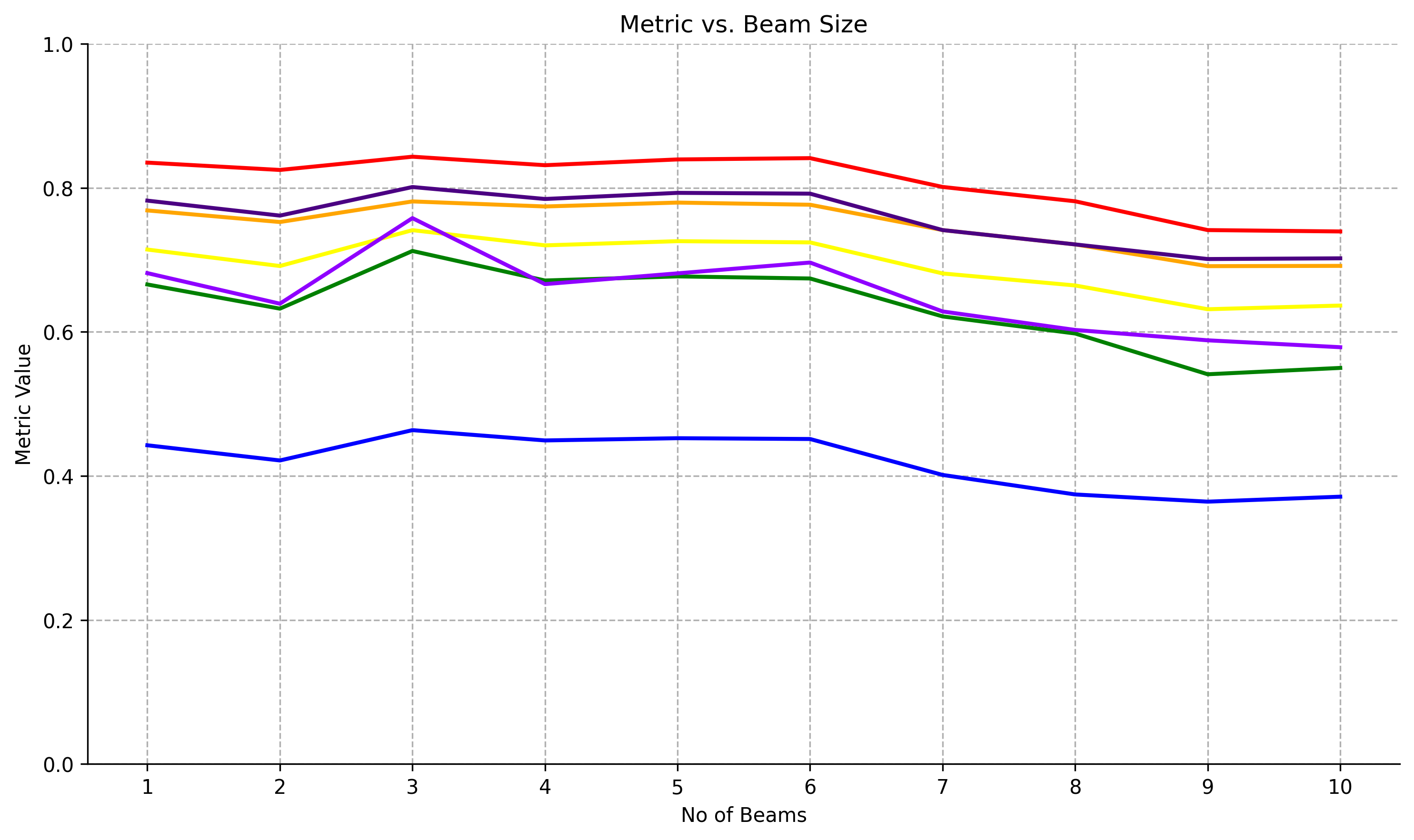}}\hfill
  \subfloat[ConvNext-RSICD\label{ConvNext-RSICD}]{\includegraphics[width=0.3\textwidth,height=125px]{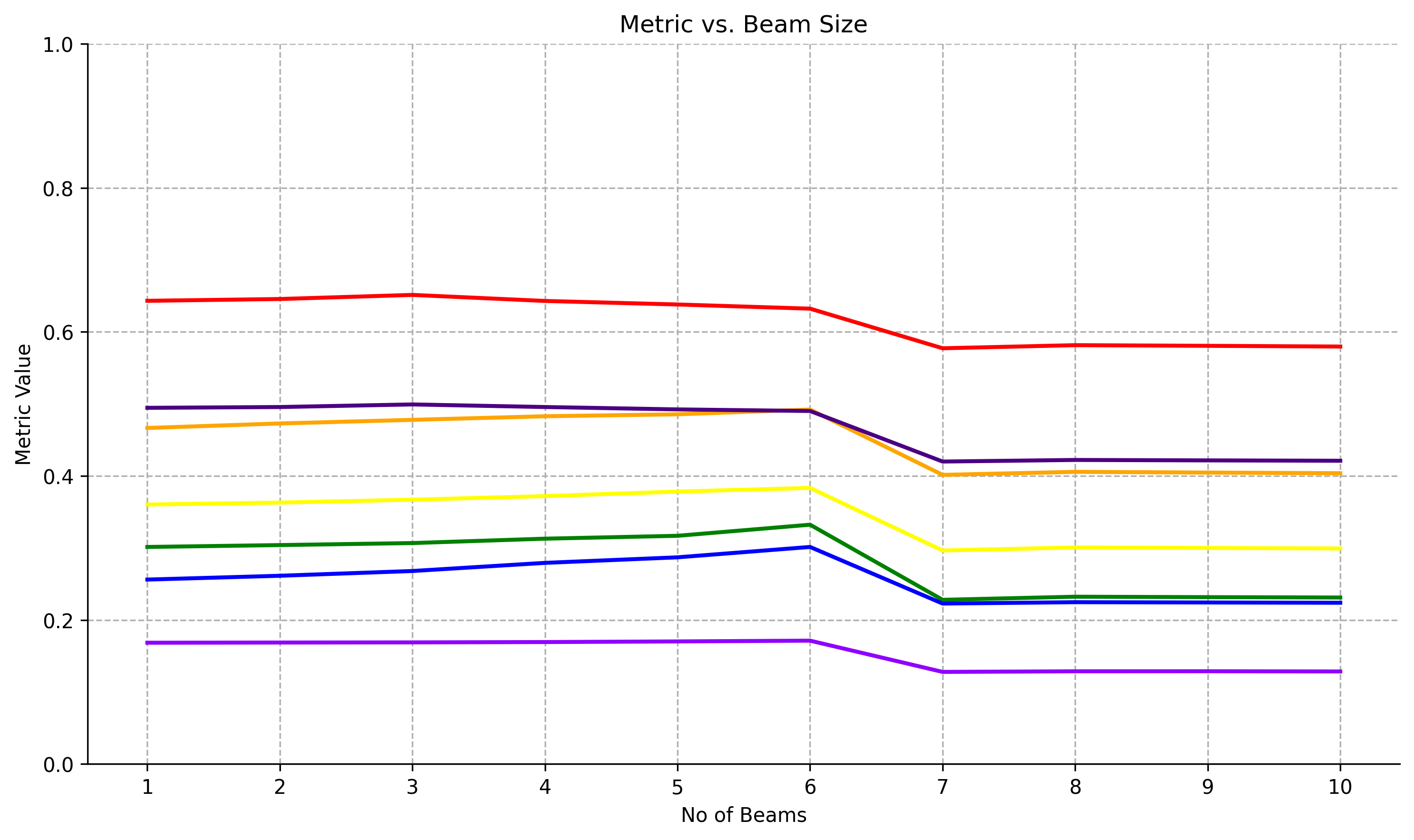}}\hfill
  \subfloat[ResNet-RSICD\label{ResNet-RSICD}]{\includegraphics[width=0.3\textwidth,height=125px]{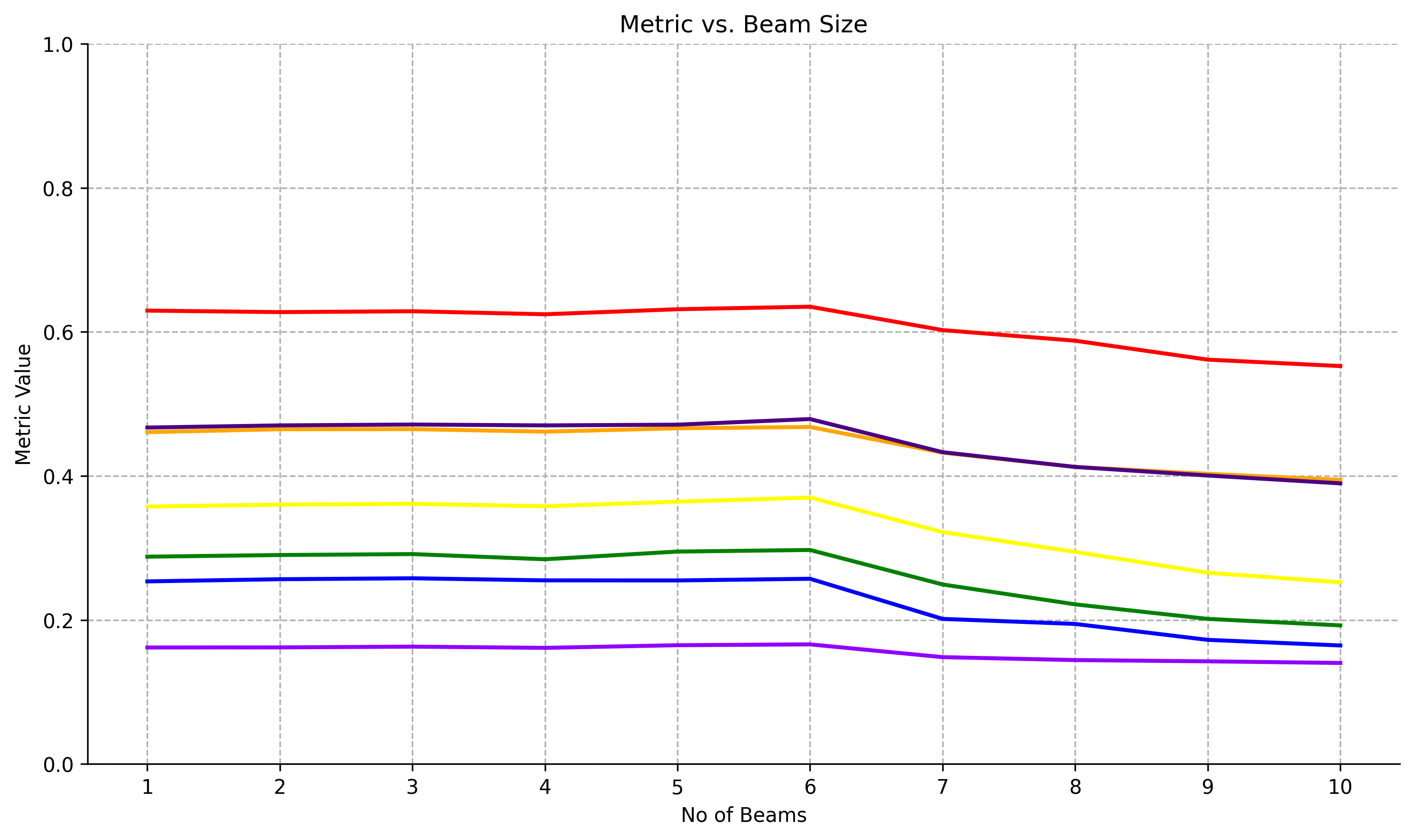}}

  \caption{
    Illustration of outputs generated with varying beam sizes across different examples. \\
    \textcolor{red}{\rule{10pt}{10pt}} BLEU-1 \quad
    \textcolor{orange}{\rule{10pt}{10pt}} BLEU-2 \quad
    \textcolor{yellow}{\rule{10pt}{10pt}} BLEU-3 \quad
    \textcolor{green}{\rule{10pt}{10pt}} BLEU-4 \quad
    \textcolor{blue}{\rule{10pt}{10pt}} METEOR \quad
    \textcolor{indigo}{\rule{10pt}{10pt}} ROUGE-L \quad
    \textcolor{violet}{\rule{10pt}{10pt}} CIDEr \quad
  }
  \label{beam_examples}
\end{figure*}
Although beam search often performs better than greedy search, prior work~\cite{cohen2019empirical} has shown that this is not always true. This is particularly evident in image captioning, where evaluation involves numerical metrics and subjective judgment. The beam search tends to favor sequences with high overall probability, which can result in overly generic captions that lack specificity. In contrast, greedy search selects the most probable word at each step without considering future possibilities. Although more limited in scope, it can sometimes produce captions that are more direct and better aligned with the image content. This is especially true for images with short and commonly generated descriptions, where beam search often leads to misclassification by overemphasizing frequent patterns.

To investigate this further, we conducted beam search experiments using beam widths ranging from $[2, 10]$ (increasing the beam width significantly increases computational time and memory requirements~\cite{hoxha2020new,cohen2019empirical}). In many practical scenarios~\cite{das2024unveiling}, particularly those that involve large-scale or time-sensitive tasks, it is not feasible to use a considerable beam width. Moreover, performance improvements beyond a certain width tend to be marginal, and the added computational cost may not be justified.

\Cref{beam_examples} illustrates the performance of the RSIC model when using ConvNext and ResNet as CNN encoders on different datasets. In this analysis, the CIDEr scores are normalized to the range $[0, 1]$ to maintain scale consistency. The results show considerable variation depending on the CNN architecture and the dataset, indicating that the optimal beam width is context-dependent. In~\Cref{ConvNext-SYDNEY}, performance improves up to beam width three and then deteriorates as beam size increases. A similar pattern is observed in~\Cref{ResNet-SYDNEY}, where performance increases until the width of the beam five and then starts to decrease. For~\Cref{ConvNext-UCM}, the performance shows an oscillating pattern up to the beam width six (which yields the highest score), followed by a downward trend. In~\Cref{ResNet-UCM}, a peculiar trend is observed in which the performance decreases at beam width two compared to one, peaks at three, and then gradually weakens. Finally, in both~\Cref{ConvNext-RSICD,ResNet-RSICD}, performance remains relatively consistent up to beam width six, with only slight differences, and then starts to taper off.
\subsection{Comparison of Different RSIC Methods}
\begin{table*}[!ht]
\centering
\caption{Performance of Different Methods on the SYDNEY Dataset}
\label{SYDNEY_results}
\resizebox{\textwidth}{!}{
\begin{tabular}{|c|c|c|c|c|c|c|c|}
\hline
Model & BLEU-1 & BLEU-2 & BLEU-3 & BLEU-4 & METEOR & ROUGE-L & CIDEr \\
\hline
CSMLF~\cite{wang2019semantic} & 0.4441 & 0.3369 & 0.2815 & 0.2408 & 0.1576 & 0.4018 & 0.9378 \\
CSMLF-FT~\cite{wang2019semantic} & 0.5998 & 0.4583 & 0.3869 & 0.3433 & 0.2475 & 0.5018 & 0.7555 \\
SVM-DBOW~\cite{hoxha2021novel} & 0.7787 & 0.6835 & 0.6023 & 0.5305 & 0.3797 & 0.6992 & 2.2722 \\
SVM-DCONC~\cite{hoxha2021novel} & 0.7547 & 0.6711 & 0.5970 & 0.5308 & 0.3643 & 0.6746 & 2.2222 \\
TrTr-CMR~\cite{wu2024trtr} & \textbf{0.8270} & \textbf{0.6994} & 0.6002 & 0.5199 & 0.3803 & 0.7220 & 2.2728 \\
R-LSTM-G~\cite{das2024unveiling} & 0.7417 & 0.6592 & 0.5925 & 0.5343 & 0.3817 & 0.6903 & 2.2032 \\
R-LSTM-B~\cite{das2024unveiling} & 0.7472 & 0.6603 & 0.5870 & 0.5254 & 0.3908 & 0.7018 & 2.1980 \\
C-MHT-G & 0.7997 & 0.6844 & \textbf{0.6325} & \textbf{0.5694} & \textbf{0.4073} & \textbf{0.7349} & \textbf{2.4945} \\
C-MHT-B & 0.7762 & 0.6742 & 0.6255 & 0.5513 & 0.3955 & 0.7281 & 2.4176 \\
\hline
\end{tabular}}
\end{table*}
\begin{table*}[!ht]
\centering
\caption{Performance of Different Methods on the UCM Dataset}
\label{UCM_results}
\resizebox{\textwidth}{!}{
\begin{tabular}{|c|c|c|c|c|c|c|c|}
\hline
Model & BLEU-1 & BLEU-2 & BLEU-3 & BLEU-4 & METEOR & ROUGE-L & CIDEr \\
\hline
CSMLF~\cite{wang2019semantic} & 0.3874 & 0.2145 & 0.1253 & 0.0915 & 0.0954 & 0.3599 & 0.3703 \\
CSMLF-FT~\cite{wang2019semantic} & 0.3671 & 0.1485 & 0.0763 & 0.0505 & 0.0944 & 0.2986 & 0.1351 \\
SVM-DBOW~\cite{hoxha2021novel} & 0.7635 & 0.6664 & 0.5869 & 0.5195 & 0.3654 & 0.6801 & 2.7142 \\
SVM-DCONC~\cite{hoxha2021novel} & 0.7653 & 0.6947 & 0.6417 & 0.5942 & 0.3702 & 0.6877 & 2.9228 \\
TrTr-CMR~\cite{wu2024trtr} & 0.8156 & 0.7091 & 0.6220 & 0.5469 & 0.3978 & 0.7442 & 2.4742 \\
R-LSTM-G~\cite{das2024unveiling} & 0.8001 & 0.7273 & 0.6675 & 0.6131 & 0.4084 & 0.7501 & 3.0616 \\
R-LSTM-B~\cite{das2024unveiling} & 0.8283 & 0.7654 & 0.7124 & 0.6621 & 0.4363 & 0.7797 & 3.2865 \\
C-MHT-G & 0.8369 & 0.7712 & 0.7143 & 0.6612 & 0.4566 & 0.8119 & 3.4582 \\
C-MHT-B & \textbf{0.8442} & \textbf{0.7802} & \textbf{0.7319} &~\textbf{0.6806} & \textbf{0.4602} & \textbf{0.8126} & \textbf{3.4953} \\
\hline
\end{tabular}}
\end{table*}
\begin{table*}[!ht]
\centering
\caption{Performance of Different Methods on the RSICD Dataset}
\label{RSICD_results}
\resizebox{\textwidth}{!}{
\begin{tabular}{|c|c|c|c|c|c|c|c|}
\hline
Model & BLEU-1 & BLEU-2 & BLEU-3 & BLEU-4 & METEOR & ROUGE-L & CIDEr \\
\hline
CSMLF~\cite{wang2019semantic} & 0.5759 & 0.3859 & 0.2832 & 0.2217 & 0.2128 & 0.4455 & 0.5297 \\
CSMLF-FT~\cite{wang2019semantic} & 0.5106 & 0.2911 & 0.1903 & 0.1352 & 0.1693 & 0.3789 & 0.3388 \\
SVM-DBOW~\cite{hoxha2021novel} & 0.6112 & 0.4277 & 0.3153 & 0.2411 & 0.2303 & 0.4588 & 0.6825 \\
SVM-DCONC~\cite{hoxha2021novel} & 0.5999 & 0.4347 & 0.3355 & 0.2689 & 0.2299 & 0.4577 & 0.6854 \\
TrTr-CMR~\cite{wu2024trtr} & 0.6201 & 0.3937 & 0.2671 & 0.1932 & 0.2399 & 0.4895 & 0.7518 \\
R-LSTM-G~\cite{das2024unveiling} & 0.6407 & 0.4676 & 0.3608 & 0.2878 & 0.2574 & 0.4745 & 0.7990 \\
R-LSTM-B~\cite{das2024unveiling} & 0.6121 & 0.4409 & 0.3385 & 0.2708 & 0.2579 & 0.4609 & 0.7594 \\
C-MHT-G & \textbf{0.6431} & 0.4665 & 0.3602 & 0.3013 & 0.2560 & 0.4945 & 0.8415 \\
C-MHT-B & 0.6380 & \textbf{0.4855} & \textbf{0.3782} & \textbf{0.3168} & \textbf{0.2869} & \textbf{0.4923} & \textbf{0.8506} \\
\hline
\end{tabular}}
\end{table*}
We present a comparison of various RSIC models in~\Cref{SYDNEY_results,UCM_results,RSICD_results}. The models evaluated include the RSIC model based on the Collective Semantic Metric Learning Framework (CSMLF)~\cite{wang2019semantic} and its variant CSMLF-FT, where FT indicates fine-tuning. The next set of models utilizes support vector machines (SVM) as decoders~\cite{hoxha2021novel}, namely SVM-DBOW and SVM-DCONC, where BOW and CONC represent two distinct sentence representations. The TrTr-CMR model~\cite{wu2024trtr} is a dual transform RSIC model based on cross-mode reasoning. It employs a Swin transformer-based encoder with a shifted window partitioning scheme for multi-scale visual feature extraction. Additionally, a Transformer language model (TLM) is designed that incorporates self- and cross-attention mechanisms of the model decoder. Finally, we consider the results of an encoder-decoder model with ResNet as the encoder and LSTM as the decoder~\cite{das2024unveiling}, with greedy search (R-LSTM-G) and beam search (R-LSTM-B) used as search techniques for caption generation.
The numerical analysis (\Cref{SYDNEY_results,UCM_results,RSICD_results}) presents the numerical results of various RSIC methods along with the ablation studies of our work. It is evident that efficient selection of the CNN and decoder improves the quality of generated captions. For the SYDNEY dataset (\Cref{SYDNEY_results}), TrTr-CMR~\cite{wu2024trtr} outperforms other models for lower-order BLEU metrics (BLEU-1 and BLEU-2), while C-MHT-G achieves the best performance for other metrics. For the UCM dataset (\Cref{UCM_results}), C-MHT-B performs the best for all metrics. Similarly, for the RSICD dataset (\Cref{RSICD_results}), C-MHT-G achieves the highest score for BLEU-1, while C-MHT-B outperforms all other models for the rest. These results indicate that ConvNext consistently delivers superior performance across all datasets, with greedy search being optimal for the SYDNEY dataset and beam search for the other two.
\subsection{Subjective Evaluation of Different CNNs}
\label{sec_subjective}
\begin{table*}[!ht]
\setlength{\tabcolsep}{0.3mm}
\footnotesize
\centering
\begin{minipage}{0.32\textwidth}
    \caption{Results of subjective evaluation on SYDNEY dataset (in \%)}
    \label{subjective_SYDNEY}
    \resizebox{\textwidth}{!}{
    \begin{tabular}{|c|c|c|c|c|}
    \hline
    Search & CNN & Rel & Part Rel & Unrel \\
    \hline
    \multirow{4}{*}{Greedy} 
     & ConvNext & 89.65 & 3.45 & 6.90 \\
     & InceptionNet & 77.59 & 8.62 & 13.79 \\
     & ResNet & 84.49 & 10.34 & 5.17 \\
     & ResNext & 86.21 & 5.17 & 8.62 \\
    \hline
    \multirow{4}{*}{Beam} 
     & ConvNext & 87.93 & 3.45 & 8.62 \\
     & InceptionNet & 79.31 & 5.17 & 15.52 \\
     & ResNet & 81.04 & 8.62 & 10.34 \\
     & VGGNet & 82.76 & 5.17 & 12.07 \\
    \hline
    \end{tabular}}
\end{minipage}
\hfill
\begin{minipage}{0.32\textwidth}
    \caption{Results of subjective evaluation on UCM dataset (in \%)}
    \label{subjective_UCM}
    \resizebox{\textwidth}{!}{
    \begin{tabular}{|c|c|c|c|c|}
    \hline
    Search & CNN & Rel & Part Rel & Unrel \\
    \hline
    \multirow{4}{*}{Greedy} 
     & ConvNext & 92.86 & 3.81 & 3.33 \\
     & DenseNet & 86.67 & 5.71 & 7.62 \\
     & ResNet & 88.09 & 2.86 & 9.05 \\
     & Wide ResNet & 87.14 & 4.29 & 8.57 \\
    \hline
    \multirow{4}{*}{Beam} 
     & ConvNext & 91.43 & 3.81 & 4.76 \\
     & DenseNet & 88.09 & 3.81 & 8.10 \\
     & ResNet & 88.10 & 5.71 & 6.19 \\
     & Wide ResNet & 89.52 & 3.81 & 6.67 \\
    \hline
    \end{tabular}}
\end{minipage}
\hfill
\begin{minipage}{0.32\textwidth}
    \caption{Results of subjective evaluation on RSICD dataset (in \%)}
    \label{subjective_RSICD}
    \resizebox{\textwidth}{!}{
    \begin{tabular}{|c|c|c|c|c|}
    \hline
    Search & CNN & Rel & Part Rel & Unrel \\
    \hline
    \multirow{4}{*}{Greedy} 
     & ConvNext & 83.63 & 6.95 & 9.42 \\
     & RegNet & 79.69 & 6.68 & 13.63 \\
     & ResNet & 80.61 & 8.23 & 11.16 \\
     & ResNext & 80.15 & 7.50 & 12.35 \\
    \hline
    \multirow{4}{*}{Beam} 
     & ConvNext & 83.71 & 6.13 & 10.16 \\
     & InceptionNet & 81.34 & 6.95 & 11.71 \\
     & MobileNetV2 & 77.95 & 8.60 & 13.45 \\
     & ResNet & 80.51 & 6.86 & 12.63 \\
    \hline
    \end{tabular}}
\end{minipage}
\end{table*}
The captioning task differs from other machine learning tasks, such as classification, where numerical analysis alone is sufficient. In tasks like classification, where the output is fixed or the number of outputs is limited, numerical analysis is often enough. However, captioning is not the same. A single image can be described by many different sentences, each of which can be equally correct. Therefore, a fixed number of test captions, with a maximum of five unique captions per image across all three datasets, is insufficient to fully evaluate the quality of the caption. In addition to numerical evaluation, subjective evaluation is also necessary~\cite{lu2017exploring,das2024unveiling}. The main purpose of subjective evaluation is to assess the generated caption from a human perspective. For this reason, we assigned the task to an annotator~\footnote{The annotator is a highly skilled professional with comprehensive training in working with RSIC models.}. Three labels were used for this evaluation, as described below:
\begin{itemize}
    \item\textbf{Related:} The generated caption accurately identifies the main object and conveys the core meaning of the image with minimal or no errors.
    \item\textbf{Partially Related:} The generated caption identifies the main object but does not fully capture the core meaning of the image or contains notable issues.
    \item\textbf{Unrelated:} The generated caption is completely unrelated to the image and misidentifies the main object.
\end{itemize}
\Cref{subjective_SYDNEY,subjective_UCM,subjective_RSICD} illustrate the subjective evaluation results for the three datasets and two search techniques. Analyzing these results, we can conclude that ConvNext performs consistently well across all scenarios. ResNet also maintains consistent performance, securing the second position in all greedy search scenario scenarios, outperforming other CNNs except ConvNext. However, the results differ for beam search. For the SYDNEY dataset, ResNet shows better performance in terms of the~\emph{Unrelated} label compared to InceptionNet and VGGNet. For the UCM dataset, Wide ResNet outperforms both ResNet and DenseNet. For the RSICD dataset, InceptionNet surpasses ResNet and MobileNetV2.
\subsection{Visual Analysis}
\begin{figure*}[!ht]
  \centering
  \subfloat[\label{example1}]{\includegraphics[width=0.25\textwidth,height=125px]{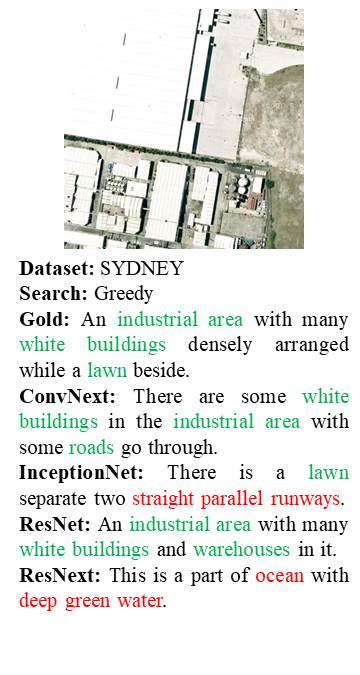}}
  \subfloat[\label{example2}]{\includegraphics[width=0.25\textwidth,height=125px]{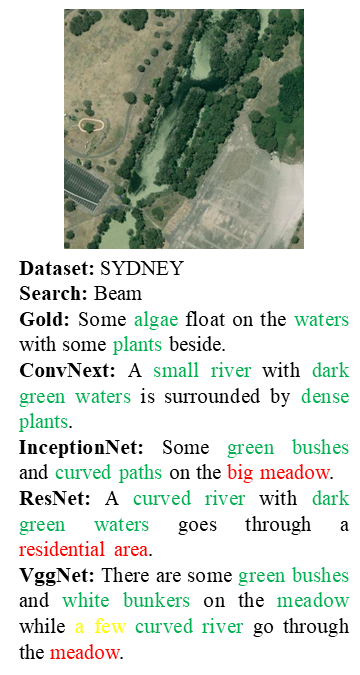}}
  \subfloat[\label{example3}]{\includegraphics[width=0.25\textwidth,height=125px]{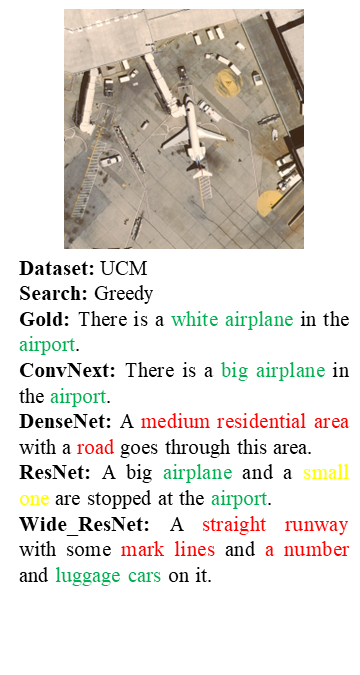}}

  \subfloat[\label{example4}]{\includegraphics[width=0.25\textwidth,height=125px]{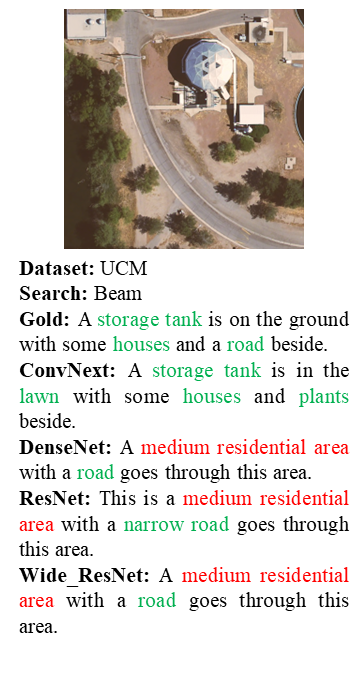}}
  \subfloat[\label{example5}]{\includegraphics[width=0.25\textwidth,height=125px]{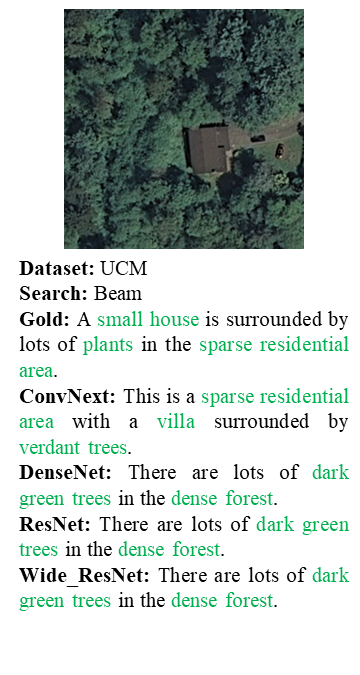}}
  \subfloat[\label{example6}]{\includegraphics[width=0.25\textwidth,height=125px]{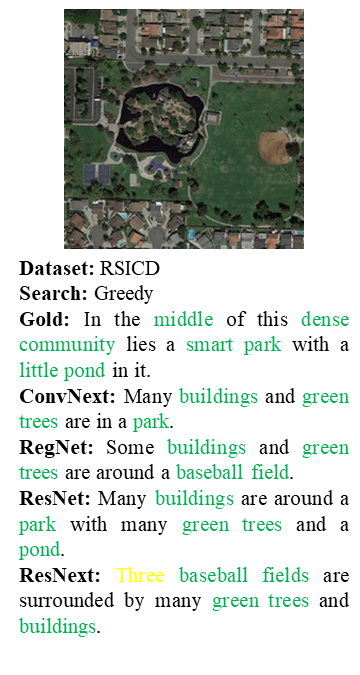}}

  \subfloat[\label{example7}]{\includegraphics[width=0.25\textwidth,height=125px]{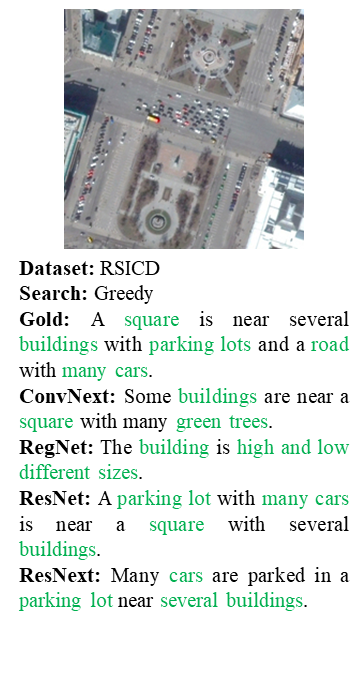}}
  \subfloat[\label{example8}]{\includegraphics[width=0.25\textwidth,height=125px]{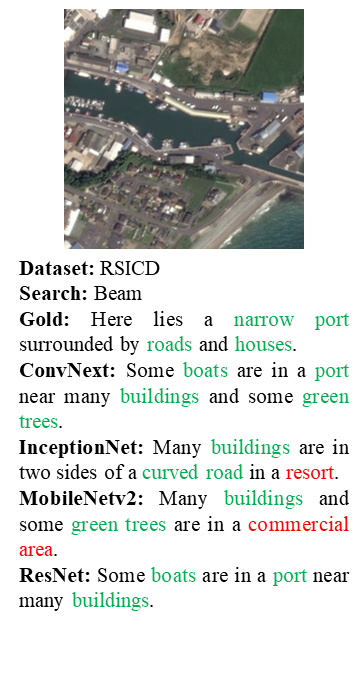}}
  \subfloat[\label{example9}]{\includegraphics[width=0.25\textwidth,height=125px]{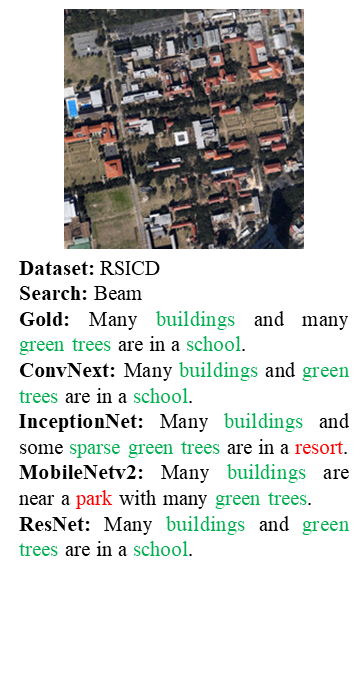}}

   \caption{Examples of RSIC on different CNNs}
   \label{examples}
\end{figure*}
\begin{figure*}[!ht]
  \centering
    \subfloat[Attention Maps Using ConvNext Encoder\label{convnext_attention}]{\includegraphics[width=0.9\textwidth,height=125px]{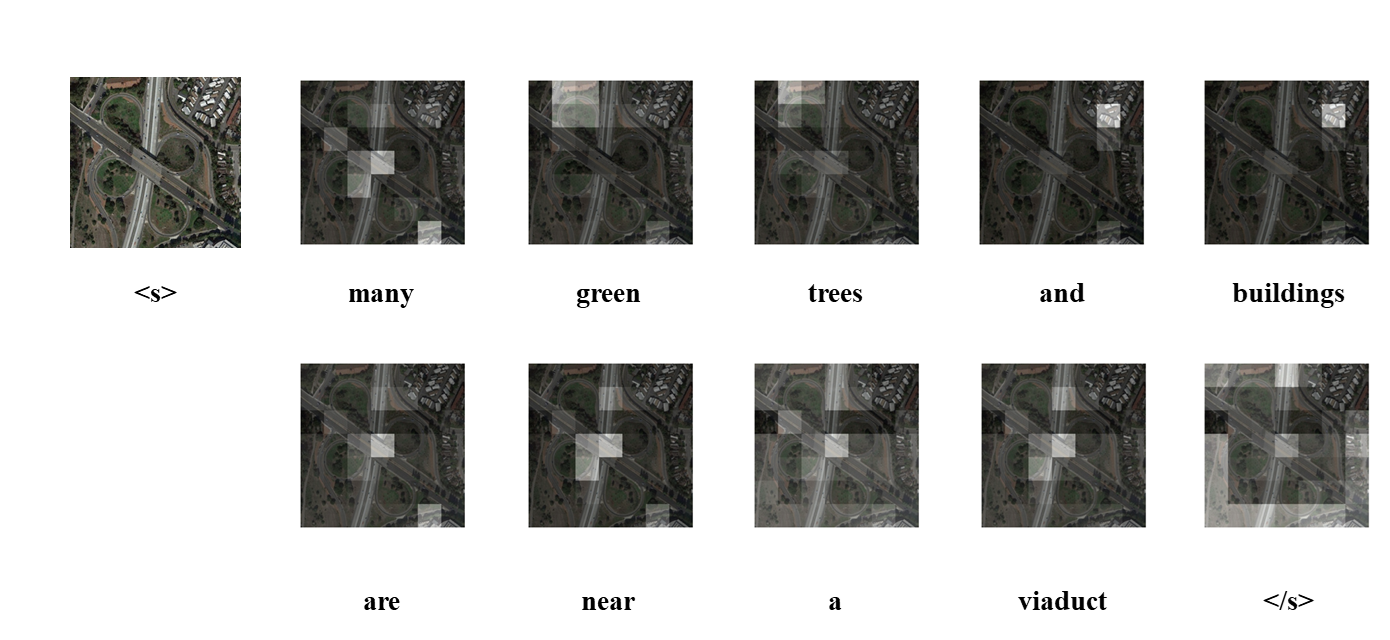}}
  
    \subfloat[Attention Maps Using ResNet Encoder\label{resnet_attention}]{\includegraphics[width=0.9\textwidth,height=125px]{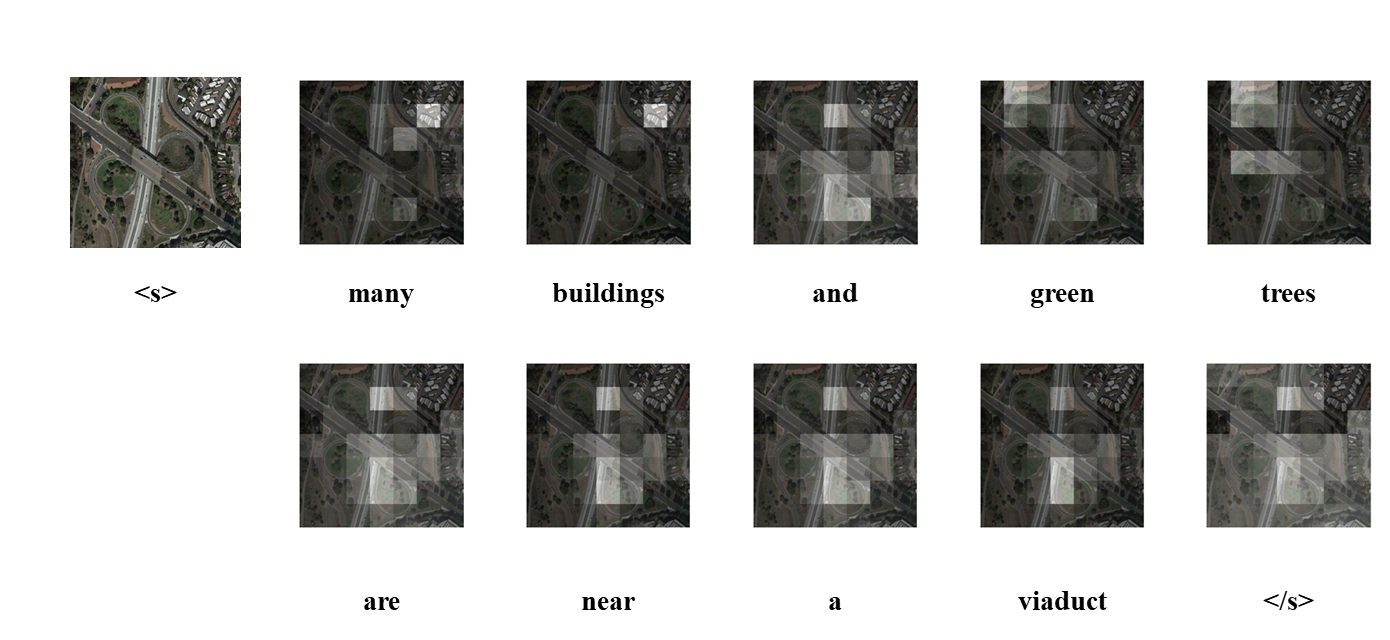}}

    \caption{Word-wise Attention Maps for Different CNN Encoders}
    \label{attention_examples}
\end{figure*}
We provide some visual examples for the CNNs in the~\emph{Good} cluster together with their captions from various datasets in~\Cref{examples}. In~\Cref{example1}, both ConvNext and ResNet effectively identify the main object~\emph{industrial area}, while InceptionNet incorrectly detects~\emph{runways} and ResNext incorrectly detects~\emph{ocean}. In~\Cref{example2}, ConvNext and ResNet correctly identify the main object~\emph{river}, while ResNet incorrectly detects~\emph{industrial area}, which is not present in the image, and both InceptionNet and VGGNet mistakenly classify it as~\emph{meadow}. In~\Cref{example3}, ConvNext and ResNet detect the primary object~\emph{airplane}, but ResNet also incorrectly identifies an additional~\emph{small airplane}, which is not present. Wide ResNet incorrectly detects~\emph{runway} and~\emph{mark lines} but detects~\emph{luggage cars}, while DenseNet completely misclassifies the image as~\emph{medium residential area}. In~\Cref{example4}, only ConvNext accurately identifies the main object~\emph{storage tank}, while DenseNet, ResNet, and Wide ResNet incorrectly classify it as~\emph{medium residential area}. In~\Cref{example5}, only ConvNext properly describes the image with the main object~\emph{villa} and the location~\emph{sparse residential area}, while DenseNet, ResNet and Wide ResNet not only fail to detect~\emph{villa} and detect the location as~\emph{forest}~\footnote{From a human perspective, both~\emph{sparse residential area} and~\emph{forest} are valid descriptions of this image; therefore, we consider both to be correct}. In~\Cref{example6}, ConvNext and ResNet successfully detect the location~\emph{park}, although ConvNext does not identify~\emph{pond}. RegNet and ResNext fail to detect~\emph{park}, but correctly identify objects such as~\emph{baseball field} (ResNext miscount is as three),~\emph{green trees}, and\emph{buildings} present in the image. In~\Cref{example7}, ConvNext and ResNet successfully identify the major object~\emph{square}, though ConvNext fails to detect~\emph{parking lot}, while RegNet and ResNext miss the main object~\emph{square}. In~\Cref{example8}, both ConvNext and ResNet accurately detect the main object~\emph{port}, while InceptionNet misclassifies it as~\emph{resort} and MobileNetV2 misclassifies it as~\emph{commercial area}. Lastly, in~\Cref{example9}, ConvNext and ResNet detect the main object~\emph{school}, while InceptionNet incorrectly classifies it as~\emph{resort} and MobileNetV2 incorrectly classifies it as~\emph{park}.

\Cref{attention_examples} illustrates how the decoder focuses on different spatial regions of the input image during the generation of each word, demonstrating the visual grounding behavior of models that use ConvNext and ResNet encoders. Here, $\left<s\right>$ marks the beginning of the token sequence, and $\left</s\right>$ indicates its end.
\subsection{Error Analysis}  
Several common errors~\cite{das2024unveiling} were identified in the three datasets when analyzing the captions generated by different CNNs. The most frequent issue is~\emph{misclassification}, which occurs when objects in an image are incorrectly identified. Some common examples include~\emph{church} and~\emph{storage tanks} being misclassified as~\emph{building},~\emph{road} being misclassified as~\emph{river} and vice versa, and~\emph{river} with~\emph{green water} being misclassified as~\emph{meadow} (\Cref{example2}).  

The second major issue is~\emph{omission of a salient object}, where the generated caption is not entirely inaccurate but does not mention the most prominent element in the image. For example, in an image depicting a~\emph{square}, peripheral objects such as~\emph{buildings} and~\emph{trees} are identified, while the~\emph{square} itself is overlooked (\Cref{example7}). Similarly, in an image of a villa in a forest, the model detects the forest but fails to identify the villa (\Cref{example5}).  

Another recurring issue is~\emph{correlation problem}, where two words frequently cooccur in the training captions. In such cases, even if only one object is present in an image, both may appear in the caption. A common example is the pair~\emph{building} and~\emph{green trees}, which are often mentioned together, even when one is absent.  

Furthermore,~\emph{counting errors} occur in images that contain multiple instances of the same object, such as~\emph{playgrounds},~\emph{storage tanks}, and~\emph{airplanes} (\Cref{example3}). The model often fails to provide an accurate count due to the lack of explicit numerical references in the training data, where descriptions usually use vague terms like~\emph{some},~\emph{several}, or~\emph{many}.  

Other errors include~\emph{incomplete captions},~\emph{failure to detect minor objects}, and~\emph{unnecessary repetition of words or fragments},~\emph{etc.}, in the generated descriptions. However, these errors occur in only a small number of captions.
\section{Conclusion}
\label{sec_conclusion}
Recent advancements in RSIC have primarily focused on the decoder, leaving the optimality of the encoder as an open question. In this work, we analyze the role of the encoder in a transformer-based RSIC model by evaluating twelve different CNN encoders from various perspectives. Our findings indicate that ConvNext consistently outperforms other CNNs in all aspects. With depthwise convolutions, layer normalization, and hierarchical feature representation, it excels in feature extraction, stability, and computational efficiency. Its strong and consistent performance across multiple tasks, including RSIC, establishes it as an optimal CNN architecture. However, the computational complexities of ConvNext are relatively high. Among the remaining models, ResNet has demonstrated stable and competitive results across different settings. In addition, ResNet has a much lower computational complexity than ConvNext. Therefore, in real-world applications where devices operate under resource constraints (such as drones, mobile devices, edge devices,~\emph{etc.}), ResNet can serve as a practical alternative. When comparing caption generation using greedy search and beam search, we observe only marginal differences across all models and datasets. The choice between the two strategies does not substantially affect the final performance, as both methods produce closely aligned results under the accumulated evaluation. In this way, this work establishes the importance of the encoder in a captioning model and provides guidance to future researchers on selecting its components effectively. In future work, we plan to explore the reduction of parameters to design some lightweight version of high-performing encoders that can preserve their representational strength while being efficient enough for deployment in resource-limited environments.
\bibliographystyle{elsarticle-num-names} 
\bibliography{mybib}

\begin{thebibliography}{48}
\expandafter\ifx\csname natexlab\endcsname\relax\def\natexlab#1{#1}\fi
\providecommand{\url}[1]{\texttt{#1}}
\providecommand{\href}[2]{#2}
\providecommand{\path}[1]{#1}
\providecommand{\DOIprefix}{doi:}
\providecommand{\ArXivprefix}{arXiv:}
\providecommand{\URLprefix}{URL: }
\providecommand{\Pubmedprefix}{pmid:}
\providecommand{\doi}[1]{\href{http://dx.doi.org/#1}{\path{#1}}}
\providecommand{\Pubmed}[1]{\href{pmid:#1}{\path{#1}}}
\providecommand{\bibinfo}[2]{#2}
\ifx\xfnm\relax \def\xfnm[#1]{\unskip,\space#1}\fi
\bibitem[{Lu et~al.(2017)Lu, Wang, Zheng, and Li}]{lu2017exploring}
\bibinfo{author}{X.~Lu}, \bibinfo{author}{B.~Wang}, \bibinfo{author}{X.~Zheng}, \bibinfo{author}{X.~Li},
\newblock \bibinfo{title}{Exploring models and data for remote sensing image caption generation},
\newblock \bibinfo{journal}{IEEE Transactions on Geoscience and Remote Sensing} \bibinfo{volume}{56} (\bibinfo{year}{2017}) \bibinfo{pages}{2183--2195}.
\bibitem[{Das and Sharma(2024)}]{das2024textgcn}
\bibinfo{author}{S.~Das}, \bibinfo{author}{R.~Sharma},
\newblock \bibinfo{title}{A textgcn-based decoding approach for improving remote sensing image captioning},
\newblock \bibinfo{journal}{IEEE Geoscience and Remote Sensing Letters}  (\bibinfo{year}{2024}).
\bibitem[{Qu et~al.(2016)Qu, Li, Tao, and Lu}]{qu2016deep}
\bibinfo{author}{B.~Qu}, \bibinfo{author}{X.~Li}, \bibinfo{author}{D.~Tao}, \bibinfo{author}{X.~Lu},
\newblock \bibinfo{title}{Deep semantic understanding of high resolution remote sensing image},
\newblock in: \bibinfo{booktitle}{2016 International conference on computer, information and telecommunication systems (Cits)}, \bibinfo{organization}{IEEE}, \bibinfo{year}{2016}, pp. \bibinfo{pages}{1--5}.
\bibitem[{Zhang et~al.(2023)Zhang, Li, Wang, Liu, Wu, Cheng, and Jiao}]{zhang2023multi}
\bibinfo{author}{X.~Zhang}, \bibinfo{author}{Y.~Li}, \bibinfo{author}{X.~Wang}, \bibinfo{author}{F.~Liu}, \bibinfo{author}{Z.~Wu}, \bibinfo{author}{X.~Cheng}, \bibinfo{author}{L.~Jiao},
\newblock \bibinfo{title}{Multi-source interactive stair attention for remote sensing image captioning},
\newblock \bibinfo{journal}{Remote Sensing} \bibinfo{volume}{15} (\bibinfo{year}{2023}) \bibinfo{pages}{579}.
\bibitem[{Li et~al.(2021)Li, Zhang, Gu, Li, Wang, Tang, and Jiao}]{li2021recurrent}
\bibinfo{author}{Y.~Li}, \bibinfo{author}{X.~Zhang}, \bibinfo{author}{J.~Gu}, \bibinfo{author}{C.~Li}, \bibinfo{author}{X.~Wang}, \bibinfo{author}{X.~Tang}, \bibinfo{author}{L.~Jiao},
\newblock \bibinfo{title}{Recurrent attention and semantic gate for remote sensing image captioning},
\newblock \bibinfo{journal}{IEEE Transactions on Geoscience and Remote Sensing} \bibinfo{volume}{60} (\bibinfo{year}{2021}) \bibinfo{pages}{1--16}.
\bibitem[{Vaswani(2017)}]{vaswani2017attention}
\bibinfo{author}{A.~Vaswani},
\newblock \bibinfo{title}{Attention is all you need},
\newblock \bibinfo{journal}{Advances in Neural Information Processing Systems}  (\bibinfo{year}{2017}).
\bibitem[{Wu et~al.(2024)Wu, Li, Jiao, Liu, Liu, and Yang}]{wu2024trtr}
\bibinfo{author}{Y.~Wu}, \bibinfo{author}{L.~Li}, \bibinfo{author}{L.~Jiao}, \bibinfo{author}{F.~Liu}, \bibinfo{author}{X.~Liu}, \bibinfo{author}{S.~Yang},
\newblock \bibinfo{title}{Trtr-cmr: Cross-modal reasoning dual transformer for remote sensing image captioning},
\newblock \bibinfo{journal}{IEEE Transactions on Geoscience and Remote Sensing}  (\bibinfo{year}{2024}).
\bibitem[{Hoxha and Melgani(2021)}]{hoxha2021novel}
\bibinfo{author}{G.~Hoxha}, \bibinfo{author}{F.~Melgani},
\newblock \bibinfo{title}{A novel svm-based decoder for remote sensing image captioning},
\newblock \bibinfo{journal}{IEEE Transactions on Geoscience and Remote Sensing} \bibinfo{volume}{60} (\bibinfo{year}{2021}) \bibinfo{pages}{1--14}.
\bibitem[{Hoxha et~al.(2020)Hoxha, Melgani, and Slaghenauffi}]{hoxha2020new}
\bibinfo{author}{G.~Hoxha}, \bibinfo{author}{F.~Melgani}, \bibinfo{author}{J.~Slaghenauffi},
\newblock \bibinfo{title}{A new cnn-rnn framework for remote sensing image captioning},
\newblock in: \bibinfo{booktitle}{2020 Mediterranean and Middle-East Geoscience and Remote Sensing Symposium (M2GARSS)}, \bibinfo{organization}{IEEE}, \bibinfo{year}{2020}, pp. \bibinfo{pages}{1--4}.
\bibitem[{Zhang et~al.(2019)Zhang, Wang, Chen, and Li}]{zhang2019multi}
\bibinfo{author}{X.~Zhang}, \bibinfo{author}{Q.~Wang}, \bibinfo{author}{S.~Chen}, \bibinfo{author}{X.~Li},
\newblock \bibinfo{title}{Multi-scale cropping mechanism for remote sensing image captioning},
\newblock in: \bibinfo{booktitle}{IGARSS 2019-2019 IEEE International Geoscience and Remote Sensing Symposium}, \bibinfo{organization}{IEEE}, \bibinfo{year}{2019}, pp. \bibinfo{pages}{10039--10042}.
\bibitem[{Li et~al.(2019)Li, Yuan, and Lu}]{li2019vision}
\bibinfo{author}{X.~Li}, \bibinfo{author}{A.~Yuan}, \bibinfo{author}{X.~Lu},
\newblock \bibinfo{title}{Vision-to-language tasks based on attributes and attention mechanism},
\newblock \bibinfo{journal}{IEEE transactions on cybernetics} \bibinfo{volume}{51} (\bibinfo{year}{2019}) \bibinfo{pages}{913--926}.
\bibitem[{Hoxha et~al.(2020)Hoxha, Melgani, and Demir}]{hoxha2020toward}
\bibinfo{author}{G.~Hoxha}, \bibinfo{author}{F.~Melgani}, \bibinfo{author}{B.~Demir},
\newblock \bibinfo{title}{Toward remote sensing image retrieval under a deep image captioning perspective},
\newblock \bibinfo{journal}{IEEE Journal of Selected Topics in Applied Earth Observations and Remote Sensing} \bibinfo{volume}{13} (\bibinfo{year}{2020}) \bibinfo{pages}{4462--4475}.
\bibitem[{Xu et~al.(2015)Xu, Ba, Kiros, Cho, Courville, Salakhudinov, Zemel, and Bengio}]{xu2015show}
\bibinfo{author}{K.~Xu}, \bibinfo{author}{J.~Ba}, \bibinfo{author}{R.~Kiros}, \bibinfo{author}{K.~Cho}, \bibinfo{author}{A.~Courville}, \bibinfo{author}{R.~Salakhudinov}, \bibinfo{author}{R.~Zemel}, \bibinfo{author}{Y.~Bengio},
\newblock \bibinfo{title}{Show, attend and tell: Neural image caption generation with visual attention},
\newblock in: \bibinfo{booktitle}{International conference on machine learning}, \bibinfo{organization}{PMLR}, \bibinfo{year}{2015}, pp. \bibinfo{pages}{2048--2057}.
\bibitem[{Sumbul et~al.(2020)Sumbul, Nayak, and Demir}]{sumbul2020sd}
\bibinfo{author}{G.~Sumbul}, \bibinfo{author}{S.~Nayak}, \bibinfo{author}{B.~Demir},
\newblock \bibinfo{title}{Sd-rsic: Summarization-driven deep remote sensing image captioning},
\newblock \bibinfo{journal}{IEEE Transactions on Geoscience and Remote Sensing} \bibinfo{volume}{59} (\bibinfo{year}{2020}) \bibinfo{pages}{6922--6934}.
\bibitem[{Wang et~al.(2022)Wang, Huang, Zhang, and Li}]{wang2022glcm}
\bibinfo{author}{Q.~Wang}, \bibinfo{author}{W.~Huang}, \bibinfo{author}{X.~Zhang}, \bibinfo{author}{X.~Li},
\newblock \bibinfo{title}{Glcm: Global--local captioning model for remote sensing image captioning},
\newblock \bibinfo{journal}{IEEE Transactions on Cybernetics} \bibinfo{volume}{53} (\bibinfo{year}{2022}) \bibinfo{pages}{6910--6922}.
\bibitem[{Liu et~al.(2022)Liu, Zhao, and Shi}]{liu2022remote}
\bibinfo{author}{C.~Liu}, \bibinfo{author}{R.~Zhao}, \bibinfo{author}{Z.~Shi},
\newblock \bibinfo{title}{Remote-sensing image captioning based on multilayer aggregated transformer},
\newblock \bibinfo{journal}{IEEE Geoscience and Remote Sensing Letters} \bibinfo{volume}{19} (\bibinfo{year}{2022}) \bibinfo{pages}{1--5}.
\bibitem[{Meng et~al.(2023)Meng, Wang, Yang, and Xiao}]{meng2023prior}
\bibinfo{author}{L.~Meng}, \bibinfo{author}{J.~Wang}, \bibinfo{author}{Y.~Yang}, \bibinfo{author}{L.~Xiao},
\newblock \bibinfo{title}{Prior knowledge-guided transformer for remote sensing image captioning},
\newblock \bibinfo{journal}{IEEE Transactions on Geoscience and Remote Sensing}  (\bibinfo{year}{2023}).
\bibitem[{Lin et~al.(2024)Lin, Wang, Ye, Wang, Yang, and Jiao}]{lin2024clip}
\bibinfo{author}{Q.~Lin}, \bibinfo{author}{S.~Wang}, \bibinfo{author}{X.~Ye}, \bibinfo{author}{R.~Wang}, \bibinfo{author}{R.~Yang}, \bibinfo{author}{L.~Jiao},
\newblock \bibinfo{title}{Clip-based grid features and masking for remote sensing image captioning},
\newblock \bibinfo{journal}{IEEE Journal of Selected Topics in Applied Earth Observations and Remote Sensing}  (\bibinfo{year}{2024}).
\bibitem[{Meng et~al.(2025)Meng, Wang, Huang, and Xiao}]{meng2025rsic}
\bibinfo{author}{L.~Meng}, \bibinfo{author}{J.~Wang}, \bibinfo{author}{Y.~Huang}, \bibinfo{author}{L.~Xiao},
\newblock \bibinfo{title}{Rsic-gmamba: A state space model with genetic operations for remote sensing image captioning},
\newblock \bibinfo{journal}{IEEE Transactions on Geoscience and Remote Sensing}  (\bibinfo{year}{2025}).
\bibitem[{Das et~al.(2024)Das, Khandelwal, and Sharma}]{das2024unveiling}
\bibinfo{author}{S.~Das}, \bibinfo{author}{A.~Khandelwal}, \bibinfo{author}{R.~Sharma},
\newblock \bibinfo{title}{Unveiling the power of convolutional neural networks: A comprehensive study on remote sensing image captioning and encoder selection},
\newblock in: \bibinfo{booktitle}{2024 International Joint Conference on Neural Networks (IJCNN)}, \bibinfo{organization}{IEEE}, \bibinfo{year}{2024}, pp. \bibinfo{pages}{1--8}.
\bibitem[{He et~al.(2016)He, Zhang, Ren, and Sun}]{he2016deep}
\bibinfo{author}{K.~He}, \bibinfo{author}{X.~Zhang}, \bibinfo{author}{S.~Ren}, \bibinfo{author}{J.~Sun},
\newblock \bibinfo{title}{Deep residual learning for image recognition},
\newblock in: \bibinfo{booktitle}{Proceedings of the IEEE conference on computer vision and pattern recognition}, \bibinfo{year}{2016}, pp. \bibinfo{pages}{770--778}.
\bibitem[{Zagoruyko(2016)}]{zagoruyko2016wide}
\bibinfo{author}{S.~Zagoruyko},
\newblock \bibinfo{title}{Wide residual networks},
\newblock \bibinfo{journal}{arXiv preprint arXiv:1605.07146}  (\bibinfo{year}{2016}).
\bibitem[{Xie et~al.(2017)Xie, Girshick, Doll{\'a}r, Tu, and He}]{xie2017aggregated}
\bibinfo{author}{S.~Xie}, \bibinfo{author}{R.~Girshick}, \bibinfo{author}{P.~Doll{\'a}r}, \bibinfo{author}{Z.~Tu}, \bibinfo{author}{K.~He},
\newblock \bibinfo{title}{Aggregated residual transformations for deep neural networks},
\newblock in: \bibinfo{booktitle}{Proceedings of the IEEE conference on computer vision and pattern recognition}, \bibinfo{year}{2017}, pp. \bibinfo{pages}{1492--1500}.
\bibitem[{Radosavovic et~al.(2020)Radosavovic, Kosaraju, Girshick, He, and Doll{\'a}r}]{radosavovic2020designing}
\bibinfo{author}{I.~Radosavovic}, \bibinfo{author}{R.~P. Kosaraju}, \bibinfo{author}{R.~Girshick}, \bibinfo{author}{K.~He}, \bibinfo{author}{P.~Doll{\'a}r},
\newblock \bibinfo{title}{Designing network design spaces},
\newblock in: \bibinfo{booktitle}{Proceedings of the IEEE/CVF conference on computer vision and pattern recognition}, \bibinfo{year}{2020}, pp. \bibinfo{pages}{10428--10436}.
\bibitem[{Simonyan and Zisserman(2014)}]{simonyan2014very}
\bibinfo{author}{K.~Simonyan}, \bibinfo{author}{A.~Zisserman},
\newblock \bibinfo{title}{Very deep convolutional networks for large-scale image recognition},
\newblock \bibinfo{journal}{arXiv preprint arXiv:1409.1556}  (\bibinfo{year}{2014}).
\bibitem[{Huang et~al.(2017)Huang, Liu, Van Der~Maaten, and Weinberger}]{huang2017densely}
\bibinfo{author}{G.~Huang}, \bibinfo{author}{Z.~Liu}, \bibinfo{author}{L.~Van Der~Maaten}, \bibinfo{author}{K.~Q. Weinberger},
\newblock \bibinfo{title}{Densely connected convolutional networks},
\newblock in: \bibinfo{booktitle}{Proceedings of the IEEE conference on computer vision and pattern recognition}, \bibinfo{year}{2017}, pp. \bibinfo{pages}{4700--4708}.
\bibitem[{Krizhevsky et~al.(2012)Krizhevsky, Sutskever, and Hinton}]{krizhevsky2012imagenet}
\bibinfo{author}{A.~Krizhevsky}, \bibinfo{author}{I.~Sutskever}, \bibinfo{author}{G.~E. Hinton},
\newblock \bibinfo{title}{Imagenet classification with deep convolutional neural networks},
\newblock \bibinfo{journal}{Advances in neural information processing systems} \bibinfo{volume}{25} (\bibinfo{year}{2012}).
\bibitem[{Szegedy et~al.(2015)Szegedy, Liu, Jia, Sermanet, Reed, Anguelov, Erhan, Vanhoucke, and Rabinovich}]{szegedy2015going}
\bibinfo{author}{C.~Szegedy}, \bibinfo{author}{W.~Liu}, \bibinfo{author}{Y.~Jia}, \bibinfo{author}{P.~Sermanet}, \bibinfo{author}{S.~Reed}, \bibinfo{author}{D.~Anguelov}, \bibinfo{author}{D.~Erhan}, \bibinfo{author}{V.~Vanhoucke}, \bibinfo{author}{A.~Rabinovich},
\newblock \bibinfo{title}{Going deeper with convolutions},
\newblock in: \bibinfo{booktitle}{Proceedings of the IEEE conference on computer vision and pattern recognition}, \bibinfo{year}{2015}, pp. \bibinfo{pages}{1--9}.
\bibitem[{Szegedy et~al.(2016)Szegedy, Vanhoucke, Ioffe, Shlens, and Wojna}]{szegedy2016rethinking}
\bibinfo{author}{C.~Szegedy}, \bibinfo{author}{V.~Vanhoucke}, \bibinfo{author}{S.~Ioffe}, \bibinfo{author}{J.~Shlens}, \bibinfo{author}{Z.~Wojna},
\newblock \bibinfo{title}{Rethinking the inception architecture for computer vision},
\newblock in: \bibinfo{booktitle}{Proceedings of the IEEE conference on computer vision and pattern recognition}, \bibinfo{year}{2016}, pp. \bibinfo{pages}{2818--2826}.
\bibitem[{Sandler et~al.(2018)Sandler, Howard, Zhu, Zhmoginov, and Chen}]{sandler2018MobileNetV2}
\bibinfo{author}{M.~Sandler}, \bibinfo{author}{A.~Howard}, \bibinfo{author}{M.~Zhu}, \bibinfo{author}{A.~Zhmoginov}, \bibinfo{author}{L.-C. Chen},
\newblock \bibinfo{title}{Mobilenetv2: Inverted residuals and linear bottlenecks},
\newblock in: \bibinfo{booktitle}{Proceedings of the IEEE conference on computer vision and pattern recognition}, \bibinfo{year}{2018}, pp. \bibinfo{pages}{4510--4520}.
\bibitem[{Howard et~al.(2019)Howard, Sandler, Chu, Chen, Chen, Tan, Wang, Zhu, Pang, Vasudevan et~al.}]{howard2019searching}
\bibinfo{author}{A.~Howard}, \bibinfo{author}{M.~Sandler}, \bibinfo{author}{G.~Chu}, \bibinfo{author}{L.-C. Chen}, \bibinfo{author}{B.~Chen}, \bibinfo{author}{M.~Tan}, \bibinfo{author}{W.~Wang}, \bibinfo{author}{Y.~Zhu}, \bibinfo{author}{R.~Pang}, \bibinfo{author}{V.~Vasudevan}, et~al.,
\newblock \bibinfo{title}{Searching for mobilenetv3},
\newblock in: \bibinfo{booktitle}{Proceedings of the IEEE/CVF international conference on computer vision}, \bibinfo{year}{2019}, pp. \bibinfo{pages}{1314--1324}.
\bibitem[{Liu et~al.(2022)Liu, Mao, Wu, Feichtenhofer, Darrell, and Xie}]{liu2022convnet}
\bibinfo{author}{Z.~Liu}, \bibinfo{author}{H.~Mao}, \bibinfo{author}{C.-Y. Wu}, \bibinfo{author}{C.~Feichtenhofer}, \bibinfo{author}{T.~Darrell}, \bibinfo{author}{S.~Xie},
\newblock \bibinfo{title}{A convnet for the 2020s},
\newblock in: \bibinfo{booktitle}{Proceedings of the IEEE/CVF conference on computer vision and pattern recognition}, \bibinfo{year}{2022}, pp. \bibinfo{pages}{11976--11986}.
\bibitem[{Zhang et~al.(2014)Zhang, Du, and Zhang}]{zhang2014saliency}
\bibinfo{author}{F.~Zhang}, \bibinfo{author}{B.~Du}, \bibinfo{author}{L.~Zhang},
\newblock \bibinfo{title}{Saliency-guided unsupervised feature learning for scene classification},
\newblock \bibinfo{journal}{IEEE Transactions on Geoscience and Remote Sensing} \bibinfo{volume}{53} (\bibinfo{year}{2014}) \bibinfo{pages}{2175--2184}.
\bibitem[{Yang and Newsam(2010)}]{yang2010bag}
\bibinfo{author}{Y.~Yang}, \bibinfo{author}{S.~Newsam},
\newblock \bibinfo{title}{Bag-of-visual-words and spatial extensions for land-use classification},
\newblock in: \bibinfo{booktitle}{Proceedings of the 18th SIGSPATIAL international conference on advances in geographic information systems}, \bibinfo{year}{2010}, pp. \bibinfo{pages}{270--279}.
\bibitem[{Xia et~al.(2017)Xia, Hu, Hu, Shi, Bai, Zhong, Zhang, and Lu}]{xia2017aid}
\bibinfo{author}{G.-S. Xia}, \bibinfo{author}{J.~Hu}, \bibinfo{author}{F.~Hu}, \bibinfo{author}{B.~Shi}, \bibinfo{author}{X.~Bai}, \bibinfo{author}{Y.~Zhong}, \bibinfo{author}{L.~Zhang}, \bibinfo{author}{X.~Lu},
\newblock \bibinfo{title}{Aid: A benchmark data set for performance evaluation of aerial scene classification},
\newblock \bibinfo{journal}{IEEE Transactions on Geoscience and Remote Sensing} \bibinfo{volume}{55} (\bibinfo{year}{2017}) \bibinfo{pages}{3965--3981}.
\bibitem[{Papineni et~al.(2002)Papineni, Roukos, Ward, and Zhu}]{papineni-etal-2002-bleu}
\bibinfo{author}{K.~Papineni}, \bibinfo{author}{S.~Roukos}, \bibinfo{author}{T.~Ward}, \bibinfo{author}{W.-J. Zhu},
\newblock \bibinfo{title}{{B}leu: a method for automatic evaluation of machine translation},
\newblock in: \bibinfo{booktitle}{Proceedings of the 40th Annual Meeting of the Association for Computational Linguistics}, \bibinfo{publisher}{Association for Computational Linguistics}, \bibinfo{address}{Philadelphia, Pennsylvania, USA}, \bibinfo{year}{2002}, pp. \bibinfo{pages}{311--318}. \URLprefix \url{https://aclanthology.org/P02-1040}. \DOIprefix\doi{10.3115/1073083.1073135}.
\bibitem[{Lavie and Agarwal(2007)}]{lavie-agarwal-2007-meteor}
\bibinfo{author}{A.~Lavie}, \bibinfo{author}{A.~Agarwal},
\newblock \bibinfo{title}{{METEOR}: An automatic metric for {MT} evaluation with high levels of correlation with human judgments},
\newblock in: \bibinfo{booktitle}{Proceedings of the Second Workshop on Statistical Machine Translation}, \bibinfo{publisher}{Association for Computational Linguistics}, \bibinfo{address}{Prague, Czech Republic}, \bibinfo{year}{2007}, pp. \bibinfo{pages}{228--231}. \URLprefix \url{https://aclanthology.org/W07-0734}.
\bibitem[{Lin(2004)}]{lin-2004-ROUGE}
\bibinfo{author}{C.-Y. Lin},
\newblock \bibinfo{title}{{ROUGE}: A package for automatic evaluation of summaries},
\newblock in: \bibinfo{booktitle}{Text Summarization Branches Out}, \bibinfo{publisher}{Association for Computational Linguistics}, \bibinfo{address}{Barcelona, Spain}, \bibinfo{year}{2004}, pp. \bibinfo{pages}{74--81}. \URLprefix \url{https://aclanthology.org/W04-1013}.
\bibitem[{Vedantam et~al.(2015)Vedantam, Lawrence~Zitnick, and Parikh}]{vedantam2015cider}
\bibinfo{author}{R.~Vedantam}, \bibinfo{author}{C.~Lawrence~Zitnick}, \bibinfo{author}{D.~Parikh},
\newblock \bibinfo{title}{Cider: Consensus-based image description evaluation},
\newblock in: \bibinfo{booktitle}{Proceedings of the IEEE conference on computer vision and pattern recognition}, \bibinfo{year}{2015}, pp. \bibinfo{pages}{4566--4575}.
\bibitem[{Dosovitskiy et~al.(2020)Dosovitskiy, Beyer, Kolesnikov, Weissenborn, Zhai, Unterthiner, Dehghani, Minderer, Heigold, Gelly et~al.}]{dosovitskiy2020image}
\bibinfo{author}{A.~Dosovitskiy}, \bibinfo{author}{L.~Beyer}, \bibinfo{author}{A.~Kolesnikov}, \bibinfo{author}{D.~Weissenborn}, \bibinfo{author}{X.~Zhai}, \bibinfo{author}{T.~Unterthiner}, \bibinfo{author}{M.~Dehghani}, \bibinfo{author}{M.~Minderer}, \bibinfo{author}{G.~Heigold}, \bibinfo{author}{S.~Gelly}, et~al.,
\newblock \bibinfo{title}{An image is worth 16x16 words: Transformers for image recognition at scale},
\newblock \bibinfo{journal}{arXiv preprint arXiv:2010.11929}  (\bibinfo{year}{2020}).
\bibitem[{Liu et~al.(2021)Liu, Lin, Cao, Hu, Wei, Zhang, Lin, and Guo}]{liu2021swin}
\bibinfo{author}{Z.~Liu}, \bibinfo{author}{Y.~Lin}, \bibinfo{author}{Y.~Cao}, \bibinfo{author}{H.~Hu}, \bibinfo{author}{Y.~Wei}, \bibinfo{author}{Z.~Zhang}, \bibinfo{author}{S.~Lin}, \bibinfo{author}{B.~Guo},
\newblock \bibinfo{title}{Swin transformer: Hierarchical vision transformer using shifted windows},
\newblock in: \bibinfo{booktitle}{Proceedings of the IEEE/CVF international conference on computer vision}, \bibinfo{year}{2021}, pp. \bibinfo{pages}{10012--10022}.
\bibitem[{Radford et~al.(2019)Radford, Wu, Child, Luan, Amodei, Sutskever et~al.}]{radford2019language}
\bibinfo{author}{A.~Radford}, \bibinfo{author}{J.~Wu}, \bibinfo{author}{R.~Child}, \bibinfo{author}{D.~Luan}, \bibinfo{author}{D.~Amodei}, \bibinfo{author}{I.~Sutskever}, et~al.,
\newblock \bibinfo{title}{Language models are unsupervised multitask learners},
\newblock \bibinfo{journal}{OpenAI blog} \bibinfo{volume}{1} (\bibinfo{year}{2019}) \bibinfo{pages}{9}.
\bibitem[{Liu et~al.(2019)Liu, Ott, Goyal, Du, Joshi, Chen, Levy, Lewis, Zettlemoyer, and Stoyanov}]{liu2019Roberta}
\bibinfo{author}{Y.~Liu}, \bibinfo{author}{M.~Ott}, \bibinfo{author}{N.~Goyal}, \bibinfo{author}{J.~Du}, \bibinfo{author}{M.~Joshi}, \bibinfo{author}{D.~Chen}, \bibinfo{author}{O.~Levy}, \bibinfo{author}{M.~Lewis}, \bibinfo{author}{L.~Zettlemoyer}, \bibinfo{author}{V.~Stoyanov},
\newblock \bibinfo{title}{Roberta: A robustly optimized bert pretraining approach},
\newblock \bibinfo{journal}{arXiv preprint arXiv:1907.11692}  (\bibinfo{year}{2019}).
\bibitem[{Lewis et~al.(2019)Lewis, Liu, Goyal, Ghazvininejad, Mohamed, Levy, Stoyanov, and Zettlemoyer}]{lewis2019bart}
\bibinfo{author}{M.~Lewis}, \bibinfo{author}{Y.~Liu}, \bibinfo{author}{N.~Goyal}, \bibinfo{author}{M.~Ghazvininejad}, \bibinfo{author}{A.~Mohamed}, \bibinfo{author}{O.~Levy}, \bibinfo{author}{V.~Stoyanov}, \bibinfo{author}{L.~Zettlemoyer},
\newblock \bibinfo{title}{Bart: Denoising sequence-to-sequence pre-training for natural language generation, translation, and comprehension},
\newblock \bibinfo{journal}{arXiv preprint arXiv:1910.13461}  (\bibinfo{year}{2019}).
\bibitem[{Gage(1994)}]{gage1994new}
\bibinfo{author}{P.~Gage},
\newblock \bibinfo{title}{A new algorithm for data compression},
\newblock \bibinfo{journal}{The C Users Journal} \bibinfo{volume}{12} (\bibinfo{year}{1994}) \bibinfo{pages}{23--38}.
\bibitem[{Sennrich et~al.(2015)Sennrich, Haddow, and Birch}]{sennrich2015neural}
\bibinfo{author}{R.~Sennrich}, \bibinfo{author}{B.~Haddow}, \bibinfo{author}{A.~Birch},
\newblock \bibinfo{title}{Neural machine translation of rare words with subword units},
\newblock \bibinfo{journal}{arXiv preprint arXiv:1508.07909}  (\bibinfo{year}{2015}).
\bibitem[{Cohen and Beck(2019)}]{cohen2019empirical}
\bibinfo{author}{E.~Cohen}, \bibinfo{author}{C.~Beck},
\newblock \bibinfo{title}{Empirical analysis of beam search performance degradation in neural sequence models},
\newblock in: \bibinfo{booktitle}{International Conference on Machine Learning}, \bibinfo{organization}{PMLR}, \bibinfo{year}{2019}, pp. \bibinfo{pages}{1290--1299}.
\bibitem[{Wang et~al.(2019)Wang, Lu, Zheng, and Li}]{wang2019semantic}
\bibinfo{author}{B.~Wang}, \bibinfo{author}{X.~Lu}, \bibinfo{author}{X.~Zheng}, \bibinfo{author}{X.~Li},
\newblock \bibinfo{title}{Semantic descriptions of high-resolution remote sensing images},
\newblock \bibinfo{journal}{IEEE Geoscience and Remote Sensing Letters} \bibinfo{volume}{16} (\bibinfo{year}{2019}) \bibinfo{pages}{1274--1278}.

\end{thebibliography}
\end{document}